\pdfoutput=1

\documentclass[11pt]{article}

\usepackage[preprint]{acl}

\usepackage{times}
\usepackage{latexsym}

\usepackage[T1]{fontenc}

\usepackage[utf8]{inputenc}

\usepackage{microtype}
\usepackage{amsmath}
\usepackage{amssymb}
\usepackage{inconsolata}

\usepackage{graphicx}

\usepackage{booktabs}
\usepackage{colortbl}
\usepackage{xcolor}
\usepackage{tabularx}
\usepackage{ragged2e} 

\usepackage{xcolor}
\usepackage[ruled,vlined]{algorithm2e}
\usepackage{caption}
\usepackage{subcaption}
\usepackage{natbib}
\usepackage{hyperref}
\usepackage{amsmath}
\usepackage{amssymb}
\usepackage{booktabs} 
\usepackage{array}
\usepackage{tikz}
\usepackage{amsmath}
\usepackage{amsfonts}
\usepackage{amssymb}
\usepackage{array} 
\usepackage{stfloats}
\usepackage{booktabs}
\usepackage{makecell}
\usepackage{mathrsfs}
\usepackage{tabularx}
\usepackage{enumitem}

\renewcommand{\phi}{\varphi}

\usepackage[export]{adjustbox}

\usepackage{booktabs}
\usepackage{hyperref}
\usepackage{longtable}
\usepackage{tcolorbox}

\usepackage{bbm}
\usepackage{stmaryrd}
\usepackage{amsmath,amssymb}

\usepackage{amsthm}

\usepackage{refcount}
\usepackage{makecell}

\newcommand{\benchmark}[0]{\texttt{ToRR}
}

\title{The Mighty \benchmark{}: \\A Benchmark for Table Reasoning and Robustness in LLMs}

\author{
    Shir Ashury-Tahan$^{\spadesuit}$$^{\heartsuit}$, 
    Yifan Mai$^{\clubsuit}$,  
    Rajmohan C$^{\spadesuit}$,  
    Ariel Gera$^{\spadesuit}$, 
    Yotam Perlitz$^{\spadesuit}$, \hfill \\
    \textbf{Asaf Yehudai$^{\spadesuit}$,
    Elron Bandel$^{\spadesuit}$,
    Leshem Choshen$^{\spadesuit\diamondsuit}$, 
    Eyal Shnarch$^{\spadesuit}$,} \hfill \\
    \textbf{Percy Liang$^{\clubsuit}$ 
    and Michal Shmueli-Scheuer$^{\spadesuit}$} \hfill \\
    $^{\spadesuit}$IBM Research, 
    $^{\heartsuit}$Bar-Ilan University, 
    $^{\clubsuit}$Stanford University, 
    $^{\diamondsuit}$MIT
    \\
    \href{mailto:shir.ashury.tahan@ibm.com}{shir.ashury.tahan@ibm.com}, 
    \href{mailto:shmueli@ibm.com}{shmueli@il.ibm.com} \hfill
}

\begin{document}
\maketitle
\begin{abstract}
Despite its real-world significance, model performance on tabular data remains underexplored, leaving uncertainty about which model to rely on and which prompt configuration to adopt.
To address this gap, we create \benchmark{}, a benchmark for Table Reasoning and Robustness, measuring model performance and robustness on table-related tasks.
The benchmark includes $10$ datasets that cover different types of table reasoning capabilities across varied domains.
\benchmark{} goes beyond model performance rankings, and is designed to reflect whether models can handle tabular data consistently and robustly,
across a variety of common table representation formats.
We present a leaderboard as well as comprehensive analyses of the results of leading models over \benchmark{}.
Our results reveal a striking pattern of brittle model behavior, where even strong models are unable to perform robustly on tabular data tasks.
We further find that no single table format consistently yields superior performance. 
However, evaluating models across multiple formats is essential for a reliable assessment of their capabilities. Moreover, we show that the reliability boost from testing multiple prompts can be equivalent to adding more test examples.
Overall, our findings show that reasoning over table tasks remains a significant challenge. The leaderboard, data and code are publicly available.

\end{abstract}


\begin{table*}
\centering
\resizebox{\textwidth}{!}{%
\small
\begin{tabular}{p{2cm} p{3cm} p{2.3cm} p{1.8cm} p{2.5cm}*{3}{>{\centering\arraybackslash}m{1.6cm}}} 
\toprule
\textbf{Dataset} & & \textbf{Task} & \textbf{Domain} & \textbf{Metric} & \textbf{Knowledge Extraction} & \textbf{Textual Reasoning} & \textbf{Numerical Reasoning} \\
\midrule
\rowcolor{white} FinQA & \citep{chen2022finqadatasetnumericalreasoning} & \RaggedRight Table QA & Finance & Program Accuracy
& \textcolor{teal}{\checkmark} & \textcolor{teal}{\checkmark} & \textcolor{teal}{\checkmark} \\
\midrule
\rowcolor{white} TableBench DA & \citep{wu2024tablebenchcomprehensivecomplexbenchmark}& \RaggedRight Data Analysis & Diverse & Rouge & \textcolor{teal}{\checkmark} & \textcolor{teal}{\checkmark} & \textcolor{teal}{\checkmark} \\
\midrule
\rowcolor{white} TableBench NR &\citep{wu2024tablebenchcomprehensivecomplexbenchmark} & \RaggedRight Table QA & Diverse & Rouge & \textcolor{teal}{\checkmark} & \textcolor{teal}{\checkmark} & \textcolor{teal}{\checkmark} \\
\midrule
\rowcolor{white} TableBench FC & \citep{wu2024tablebenchcomprehensivecomplexbenchmark}& \RaggedRight Table QA & Diverse & Rouge & \textcolor{teal}{\checkmark} & \textcolor{teal}{\checkmark}  & \textcolor{orange}{\textasciitilde}  \\
\midrule
\rowcolor{white} WikiTQ & \citep{pasupat-liang-2015-compositional}& \RaggedRight Table QA & Wikipedia & F1 Strings & \textcolor{teal}{\checkmark} & \textcolor{teal}{\checkmark}  & \textcolor{orange}{\textasciitilde}  \\
\midrule
\rowcolor{white} TabFact & \citep{chen2020tabfactlargescaledatasettablebased}& \RaggedRight Fact Verification & Wikipedia & Accuracy & \textcolor{teal}{\checkmark} & \textcolor{teal}{\checkmark}  & \textcolor{orange}{\textasciitilde}  \\
\midrule
\rowcolor{white} QTSumm &\citep{zhao2023qtsummqueryfocusedsummarizationtabular} & \RaggedRight Table-to-Text QA& Wikipedia & Rouge & \textcolor{teal}{\checkmark} & \textcolor{orange}{\textasciitilde} & \textcolor{orange}{\textasciitilde} \\
\midrule
\rowcolor{white} SciGen &\citep{moosavi2021scigen} & \RaggedRight Table-to-Text & Science & Rouge & \textcolor{orange}{\textasciitilde} & \textcolor{purple}{\boldmath$\times$} & \textcolor{orange}{\textasciitilde} \\
\midrule
\rowcolor{white} NumericNLG & \citep{suadaa-etal-2021-towards} & \RaggedRight Table-to-Text & Science & Rouge & \textcolor{orange}{\textasciitilde} & \textcolor{purple}{\boldmath$\times$} & \textcolor{orange}{\textasciitilde} \\
\midrule
\rowcolor{white} TURL CTA &\citep{deng2020turltableunderstandingrepresentation} & \RaggedRight Classification & Wikipedia & Exact Match & \textcolor{orange}{\textasciitilde} & \textcolor{purple}{\boldmath$\times$} & \textcolor{purple}{\boldmath$\times$} \\
\bottomrule
\end{tabular}
}
\parbox{\linewidth}{\small
\begin{tabular}{p{2.5cm} p{1.5cm} p{1.8cm} p{2.5cm} p{2.2cm}}
   \\
   & \textbf{Legend:} & Required \textcolor{teal}{\checkmark} &  Partially Required \textcolor{orange}{\textasciitilde} & Not Required \textcolor{purple}{\boldmath$\times$}\\
\end{tabular}
}

\caption{The selected datasets for \benchmark{} along with their properties. The $3$ columns on the right reflect the required skills to solve each dataset, based on our analysis (\S\ref{sec:bench_construct}). Further details in Appendix~\ref{appendix:benchmark_details}.
}
\label{tab:datasets_final_colored}

\end{table*}

\section{Introduction}\label{intro}

Tabular data are ubiquitous across real-world use cases and tasks. Hence, the ability to understand and process tables is a crucial skill for Large Language Models (LLMs). Tabular processing capabilities can manifest in a wide range of NLP tasks, including table-to-text~\citep{moosavi2021scigen,suadaa-etal-2021-towards}, table question answering~\citep{pasupat-liang-2015-compositional,wu2024tablebenchcomprehensivecomplexbenchmark} and table fact verification \citep{chen2020tabfactlargescaledatasettablebased,gu2022pasta}. In order to solve such problems, LLMs must correctly parse tabular structures, but must also apply various levels of textual and numerical reasoning over the table contents. Thus, tabular tasks are a challenging test of LLMs' capabilities and practical utility.

\begin{figure}[]

  \centering
  \includegraphics[width=0.45\textwidth]{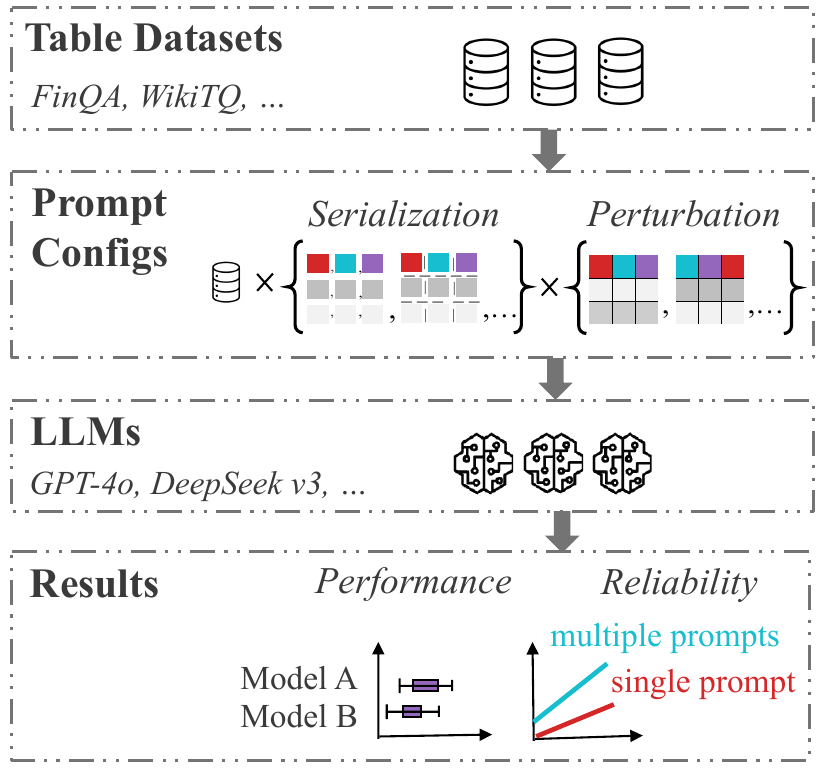}
  \caption{Overview of \benchmark{}. 
    We evaluate LLMs on several tabular reasoning datasets using a variety of prompt configurations. Each configuration includes a table serialization — a method for converting the table into a string format (e.g., HTML serialization) — and may include a perturbation to the table structure (e.g., shuffling row order).
    Our results explore \textit{model performance} and the effects of \textit{prompt variability}. Our analysis demonstrates that for any number of examples, testing more prompt configurations increases the evaluation reliability.
    }
    \label{fig:schema}
\end{figure}

Although prior studies have investigated LLM performance on tabular tasks~\citep{ruan2024language, fang2024large, lu2024large, chen2022large}, their scope remains limited, typically constrained to specific tasks, narrow domains, or a small subset of models. Consequently, these evaluations fail to provide a comprehensive understanding of LLM capabilities in structured data reasoning. Furthermore, they fail to account for the complexities of real-world applications, as they do not encompass the full range of skills required for working with tables. Importantly, they do not systematically assess the \textit{robustness} of LLMs to variations in table formatting.

Given that real-world tables frequently appear in diverse yet semantically equivalent textual formats~\citep{singha2023tabular, sui2024table, zhao2023robut, bhandari2024robustness}, it is crucial to assess LLM capabilities across formats. Such evaluation further provides insight into the models’ internal representations and their ability to generalize beyond specific formatting styles.

In this work, we paint a comprehensive picture of the ability of LLMs on downstream table understanding and table reasoning tasks. To this end, we design a pipeline for evaluating LLMs on tabular tasks, as illustrated in Figure~\ref{fig:schema}, and collect $10$ datasets belonging to $6$ diverse tabular tasks, from multiple domains. Those together amount to our benchmark, \benchmark{}, a broad coverage benchmark, testing different levels of table skills. 

The focus on robustness is inherent to the design of \benchmark{}. Our benchmark examines how models respond to differing prompt configurations -- table serialization formats, as well as table perturbations. Thus, going beyond bottom-line model rankings and performance, we are able to conduct an in-depth analysis of LLM behavior on tabular tasks.  

We evaluate the performance and robustness of $14$ leading LLMs spanning $7$ model families on \benchmark{}, which highlights significant gaps in model capabilities, even for industry-leading LLMs.

Our results demonstrate that existing models struggle with tasks that require table understanding and reasoning, and exhibit brittle and inconsistent response patterns. At the same time, we find that no single table format consistently yields superior performance.

We further analyze the broader implications of our benchmark results, focusing on the aspect of accounting for model robustness. Our meta-evaluation analysis highlights the importance of incorporating a large number of prompt configurations into the evaluation process, demonstrating the benefits of the \benchmark{} benchmark design for evaluation reliability compared to existing benchmarks.

We start by describing the \benchmark{} benchmark design (\S\ref{sec:bench_construct}), followed by the results and model analysis (\S\ref{section:results}).
We then discuss the validity of the benchmark in Section~\ref{subsec:benchmark-characteristics}. Finally, Section~\ref{section:prompt_impact_on_eval} analyzes robustness using \benchmark{} data, highlighting behaviors with practical implications beyond tabular reasoning.

Our main contributions are as follows:

\begin{enumerate}[leftmargin=10pt]
    \item We introduce a comprehensive benchmark
    of downstream tabular data tasks, encompassing diverse tasks and incorporating model robustness measurements (\S\ref{sec:bench_construct}).
    \item We present evaluation results over $14$ LLMs. Our study uncovers substantial limitations in the tabular capabilities of existing LLMs, challenging assumptions about their generalization power in structured data contexts (\S\ref{subsec:  capabilities}, \S\ref{subsec: perform_aspects}).
    \item We demonstrate that no single table format consistently yields better model performance -- an observation with implications for both model evaluation and practical use (\S\ref{sec: prompt_performance}). 
    \item In a meta-evaluation analysis, we show that incorporating prompt configuration into the evaluation process \textit{consistently} enhances model evaluation reliability and can help compensate for smaller test sets (\S\ref{section:prompt_impact_on_eval}).
    \item We release the leaderboard, complete model inference outputs and the code.
\end{enumerate}

\section{\benchmark{} Construction} \label{sec:bench_construct}

We reviewed numerous existing datasets for downstream tabular data tasks, prioritizing \textit{challenging} ones based on the required reasoning abilities.
Also, we opted for datasets where textual tables can be \textit{directly incorporated} into the prompt\footnote{We use tables that can be serialized into text and fit within the context of five demos.}, eliminating the need for external tools (e.g., retrieval, SQL queries, agents).
Table~\ref{tab:datasets_final_colored} presents the selected datasets and their attributes, which were analyzed manually.
As can be seen, the datasets are diverse in both the target task and domain. Further details on the selected and additional datasets, as well as evaluation metrics, are provided in Appendix~\ref{appendix:benchmark_details}.

For a better understanding of the skills needed to solve each dataset, we performed a qualitative manual analysis of the nature of the tasks. Specifically, we identify \textit{$3$ key skills} ranked from easiest to hardest: 

\begin{enumerate}[leftmargin=0pt]
    \item \textbf{Knowledge Extraction} – Extraction of relevant information from the table, such as specific fields, relations, or entities (e.g., \textit{"What was the only year Keene won Class AA?"}; WikiTQ).
    
    \item \textbf{Textual Reasoning} – Deducing conclusions by combining the accompanying text with the data contained in the table (e.g., \textit{"Considering the weighted average fair value of options, what was the change of shares vested from 2005 to 2006?"}; FinQA).
    
    \item \textbf{Numerical Reasoning} – Performing calculations on the table, such as aggregating information from multiple cells (e.g., \textit{"What is the total GDP of all South American countries listed in the table according to the 2011 IMF estimates?"}; TableBench).
\end{enumerate}

As shown in Table~\ref{tab:datasets_final_colored}, knowledge extraction ability is a key requirement across all datasets. The level of textual reasoning and numerical reasoning required over the tables varies across datasets and tasks.
It also demonstrates that our selection of datasets in \benchmark{} covers a range of challenge levels. This will be further supported by a correlation with our robustness analysis (see \S\ref{section:prompt_impact_on_eval}).

\subsection{Prompt Configurations} \label{sec:table_forms}
In real-world tabular tasks, the tables provided as input to an LLM can be represented in different formats; for instance, in JSON or HTML format.
Although these formats encode the same content (column names, cell values etc.), LLMs may perform differently depending on the input format.
Thus, these formats offer insight into how models represent tables and, specifically, on the extent to which they are able to generalize table understanding and reasoning across different formats.

To this end, we manipulate the table format across $2$ dimensions. First, for each input table, we examine $7$ \textbf{serializations} i.e., methods to represent the contents of the table as a string (e.g., using \textit{JSON} or \textit{HTML}). In addition, we explore $4$ structural \textbf{perturbations} which are applied to the tables; for example, shuffling the order of rows, or transposing the rows and columns. An example of the resulting prompts is shown in Figure~\ref{fig:prompts_example}. For details on all prompt configurations, see Appendix~\ref{appendix:table_forms_details}, and for an explanation of why this helps in addressing contamination is provided, refer to Appendix~\ref{app:sec_contamination}.

\subsection{Metrics} \label{subsec:metrics}
\benchmark{} consists of datasets denoted as $D$, and each dataset $d \in D$ contains examples $\{(x_i, y_i)\}_{i \in d}$, where $x_i$ represents the input, and $y_i$ is the ground-truth response. 
Each input can be represented using one of $c \in C$ prompt configurations (\S\ref{sec:table_forms}), denoted as $x_i^c$.
Each dataset in \benchmark{} is associated with a specific evaluation metric (see Table~\ref{tab:datasets_final_colored}). All metrics fall within the range $[0.0 , 1.0]$, ensuring comparability of aggregated scores across datasets. We denote the score function for an example, as defined by the dataset, as $S$.


\textbf{Model Performance}
The performance of a model $M$ is its ability to solve table-related tasks, 
regardless of the table format.
Let $M(x_i^c)$ denote the output of model $M$ for input $x_i^c$. 
The performance score, $\mathcal{P}$, is the average across prompts and is defined as:
\[
\mathcal{P}_{M} = \frac{1}{|D|} \sum_{d \in D} \frac{1}{|d|} \sum_{i\in d} \frac{1}{|C|} \sum_{c \in C} S(M(x_i^c), y_i)
\]

\textbf{Model Robustness}
A robust model is expected to perform similarly on different prompt configurations of the same example, i.e., to have a low variance over the example performance scores.
Thus, we define the robustness score, $\mathcal{R}$, as the complement of the average score range per example:
\begin{align*}
\mathcal{R}_{M}  = 1 - \frac{1}{|D|} \sum_{d \in D} \frac{1}{|d|} \sum_{i\in d}  \bigg[&\max_{c\in C}S(M(x_i^c), y_i) \\ - &\min_{c\in C}S(M(x_i^c), y_i) \bigg]
\end{align*}

\begin{figure*}
    \centering
    \includegraphics[width=0.9\linewidth]{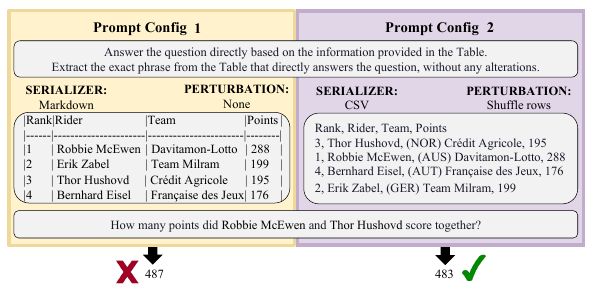}
    \caption{
    \benchmark{} prompt examples with identical instructions but different table formats: markdown (left) and shuffled CSV (right) (\S\ref{sec:table_forms}).
    While all prompts convey the same information and require the exact same answer, even state-of-the-art models struggle with solving them consistently (\S\ref{section:results}).
    }
    \label{fig:prompts_example}
\end{figure*}

\subsection{Setup} \label{subsec:  Setup}
For each dataset, we sample $100$ examples from the test set. For each example, as mentioned in \S\ref{sec:table_forms}, we represent its table using $7$ different serializations. We also apply $4$ different structural perturbations. As the perturbations are orthogonal to the chosen serialization, this yields a total of $35$ $p \in P$ prompt configurations ($7$ serializations $\times$ $4$ perturbations + $7$ without perturbation). 

We run a total of $14$ models over \benchmark{}, utilizing the Together AI\footnote{\href{https://www.together.ai/}{https://www.together.ai/}} inference engine. We use 5-shot prompting\footnote{Except WikiTQ, which we truncated to one-shot due to length.}, where each set of 5 shots is randomly sampled for each example, with greedy decoding and limit the maximum token output to $512$. Each model was run on the same set of $100$ examples per dataset\footnote{TableBench FC has 96 examples; we used 79 for \benchmark{}, the rest as demonstrations.} $\times$ $35$ prompt configurations.

We utilize the \texttt{Unitxt} library \citep{bandel-etal-2024-unitxt} to ensure that \benchmark{} is shareable and reproducible, as also highlighted by \citet{reuel2024betterbenchassessingaibenchmarks}. The modular customization of the library allowed us to manipulate the choice of table serialization and apply perturbations while keeping other aspects of prompt design (e.g., few-shot examples) constant. 
Further details about usage can be found in Appendix~\ref{appendix:usage_options}.

\section{Results \& Analysis} \label{section:results}
\benchmark{} offers multiple insights into the capabilities and performance of models on tabular tasks.

We present the high-level results of \benchmark{} in \S\ref{subsec:  capabilities}. 
Next, we analyze advanced aspects of model performance in \S\ref{subsec: perform_aspects}.
Finally, we examine how prompt configurations affect performance
(\S\ref{subsec: eval_serializer}).

\begin{table*}
\centering
\resizebox{\textwidth}{!}{%

\begin{tabular}{l|cc|cccccccccc}
Model & $\mathcal{P}$ ($\uparrow$) & $\mathcal{R}$ ($\uparrow$) & FinQA & \makecell{Numeric-\\NLG} & QTSumm & SciGen &\makecell{Tab \\ Fact }& \makecell{TableBench \\ DA} & \makecell{TableBench \\FC} & \makecell{TableBench \\NR} &  \makecell{TURL\\ CTA} & WikiTQ \\

\midrule
claude-3-5-sonnet & \textbf{.50} & \textbf{.70} & .43 & .17 & .37 & .15 & \textbf{.86} & .31 & .70 & \textbf{.42} & \textbf{.68} & \textbf{.92}  \\
claude-3-5-haiku & .43 & .61 & .35 & .17 & .34 & .15 & .79 & .23 & .63 & .20 & .55 & .85 \\
\specialrule{0.1pt}{1pt}{1pt}
gpt-4o & \textbf{.50} & \underline{.69} & .40 & \textbf{.19} & \textbf{.43} & \textbf{.17} & \underline{.83} & \textbf{.33} & \textbf{.73} & \underline{.40} & .60 & \underline{.91}  \\
gpt-4o-mini & .43 & .62 & .34 & \underline{.18} & .40 & \underline{.16} & .65 & .27 & .64 & .24 & .54 & .88  \\
\specialrule{0.1pt}{1pt}{1pt}
deepseek-v3 & \textbf{.50} & .67 & \underline{.46} & \textbf{.19} & .41 & \underline{.16} & .81 & \underline{.32} & \underline{.71} & .35 & \underline{.66} & \underline{.91}   \\
\specialrule{0.1pt}{1pt}{1pt}
gemini-1.5-pro & \underline{.48} & .65 & \textbf{.47} & \textbf{.19} & .38 & \underline{.16} & .80 & .29 & .70 & .32 & .62 & .89 \\
gemini-1.5-flash & .45 & .64 & .43 & \textbf{.19} & .34 & \underline{.16} & .77 & .28 & .70 & .21 & .57 & .88 \\
\specialrule{0.1pt}{1pt}{1pt}
qwen2-72b-i & .45 & .63 & .38 & .17 & \underline{.42} & .15 & .78 & .27 & .68 & .23 & .60 & .86  \\
\specialrule{0.1pt}{1pt}{1pt}
llama-3-1-405b-i & .46 & .60 & .36 & .12 & \underline{.42} & .11 & .82 & .30 & .65 & .31 & .62 & .90 \\
llama-3-1-70b-i & .44 & .59 & .37 & .12 & \underline{.42} & .10 & .73 & .30 & .63 & .27 & .60 & \underline{.91}  \\
llama-3-1-8b-i & .29 & .53 & .04 & .11 & .35 & .09 & .55 & .16 & .52 & .10 & .18 & .80   \\
\specialrule{0.1pt}{1pt}{1pt}
mixtral-8x22b-i & .41 & .58 & .28 & \textbf{.19} & .41 & \textbf{.17} & .74  & .24 & .68 & .20 & .54 & .67  \\
mixtral-8x7b-i & .35 & .49 & .20 & \underline{.18} & .36 & \textbf{.17} & .65 & .23 & .59 & .12 & .35 & .62  \\
mistral-7b-i & .32 & .54 & .19 & .14 & .38 & .13 & .56 & .22 & .55 & .10 & .30 & .68  
\end{tabular}
}
\caption{
    Main results of LLMs on \benchmark{}: overall performance ($\mathcal{P}$) and robustness ($\mathcal{R}$) scores, along with performance scores for each dataset.
    Best scores are marked with bold, second best are underlined. }
    \label{table:perform}
\end{table*}

\begin{figure}[h]
  \centering
  \includegraphics[width=0.5\textwidth]{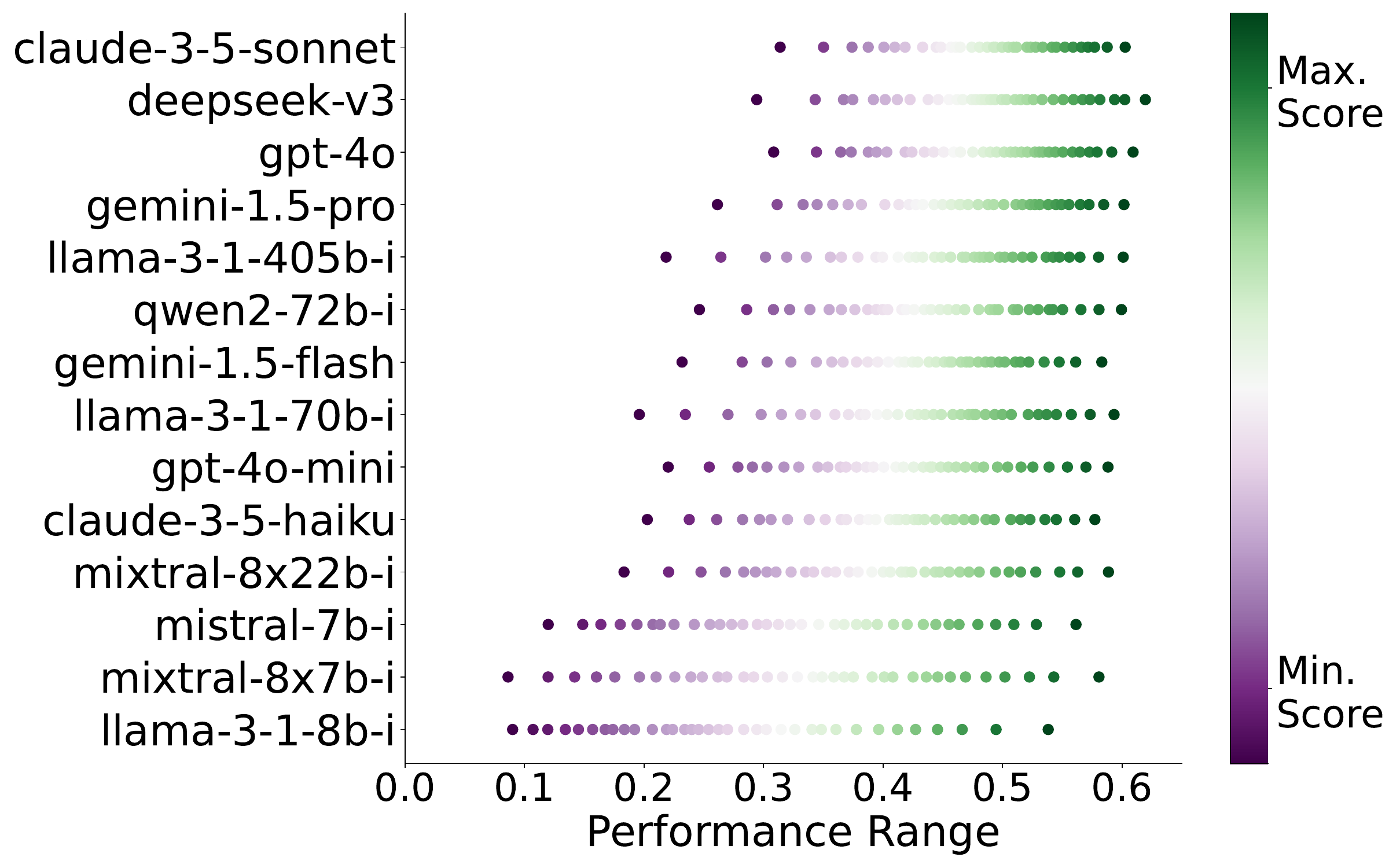}
    \caption{
    For each example we obtained $35$ performance scores using different prompt configurations. The example scores are assigned an index in the range $[1, 35]$, ordered from lowest to highest performance. The plot depicts an average aggregation of each index across all examples.
    Models exhibit a wide range of scores, reflecting low robustness. 
    }
    \label{fig_perform}
\end{figure}

\subsection{Model Capabilities} \label{subsec:  capabilities}

Table \ref{table:perform} showcases the main results of $14$ open and closed models on \benchmark{}. \textit{claude-3-5-sonnet, gpt-4o} and \textit{deepseek-v3} 
(full model names are in Appendix \ref{appendix:model_names})
outperform others in most of the datasets, however models within the larger size range demonstrate similar performance. These models also have higher robustness scores, and overall, it appears that robustness scores correlate with performance scores, e.g., models with better performance tend to be more robust. Two main trends that \benchmark{} show are:


\paragraph{Current models exhibit limitations in processing tabular data}

The absolute scores of all models are medium-low, at best reaching $0.50$ (see Table \ref{table:perform}); but also, the gap between lower-performing models and higher-performing models is narrow. This indicates that better models are only moderately better at table understanding. 
Moreover, paired Cohen's $d$, an effect size metric, shows most model comparisons have small, often non-significant, practical differences. See Appendix~\ref{appendix:statistical_testing} for more details.

\paragraph{All models are not robust.}

For each model and example from the selected datasets, we obtain a set of $35$ scores (\S\ref{subsec:  Setup}) that should all reflect how well the model handles this example and hence expected to mainly agree with each other. Yet in practice, these scores vary widely. 

Figure~\ref{fig_perform} illustrates this range of scores for each model, aggregated across examples. As can be seen, the minimal score and the maximal score yield entirely different estimates of model performance. 
Thus, we see that the models exhibit strikingly brittle behavior, and are highly influenced by the choice of configuration. This suggests that current models do not have a stable and generalizable table representation that persists across table formats. 


\subsection{Model Performance Trends} \label{subsec: perform_aspects}

As can be seen in Table~\ref{table:perform}, performance within model families is directly correlated to the model size. 
However, the differences in scores within families tend to be relatively small, and are $0.07$ on average\footnote{Averaging over the differences between pairs of models within the same family.}. 
Across model families, size does not always indicate performance; for example, \textit{qwen2-72b-instruct} outperforms \textit{llama-3.1-405b-instruct} in both performance and robustness.

Table~\ref{table:perform} compares model behavior across datasets, highlighting the strengths and weaknesses of each model.
For example, while \textit{claude-3-5-sonnet} outperforms others in classification tasks (e.g., TabFact), \textit{mixtral-8x22b-instruct} performs the worst in them. However, \textit{mixtral-8x22b-instruct} shows better capabilities than \textit{claude-3-5-sonnet} in Table-to-Text tasks (e.g., QTSumm).

The order of models in Table~\ref{table:perform} reflects an overall advantage of closed models over open models. 
Delving deeper, it appears they outperform in all tasks we examined. Additional analysis can be found in Appendix~\ref{app:perform_analysis}.

\subsection{Performance by Prompt Configuration} \label{sec: prompt_performance}

The notable lack of model robustness observed above raises the question of whether it results from certain configurations outperforming others.

\paragraph{No serializer leads to superior performance.} \label{subsec: eval_serializer}

To evaluate whether specific serializations give rise to better model performance, we calculate the win-rate of serializers at the example level. Then, we aggregate the results for all models and serializers and find that no serializer consistently outperforms others.

When breaking down the results by model, we do find some weak effects of preferred serializations for specific models, with a maximum difference of $0.06$ in overall model performance across serializers.
Additional details and figures can be found in Appendix~\ref{app:analysis_sec:serializers}.


\paragraph{Perturbations have no consistent effect.} \label{subsec: eval_perturb}

We find that the perturbations outlined in \S\ref{sec:table_forms} do not consistently affect performance, neither decreasing nor increasing model performance. For further details, refer to Appendix~\ref{app:analysis_sec:perturbs}.

\section{Properties of \benchmark{}} \label{subsec:benchmark-characteristics}

The reliability and validity of a benchmark are critical. We assess \benchmark{} properties through statistical testing, separability analysis, and dataset agreement.

\paragraph{Statistical Testing}

Tied results in Section~\ref{section:results} raise concerns about the benchmark’s ability to distinguish between models. Following \citet{ackerman2025statisticalmultimetricevaluationvisualization}, we perform significance tests on all pairwise model comparisons across datasets. Many comparisons show statistically significant differences.\footnote{Differences between p-values and effect sizes (see \S\ref{subsec:  capabilities}) may stem from the large sample size in \benchmark{}. Details in Appendix~\ref{appendix:statistical_testing}.}

\begin{figure*}[t]
    \begin{subfigure}{0.4\textwidth}
        \centering
        \includegraphics[width=\textwidth]{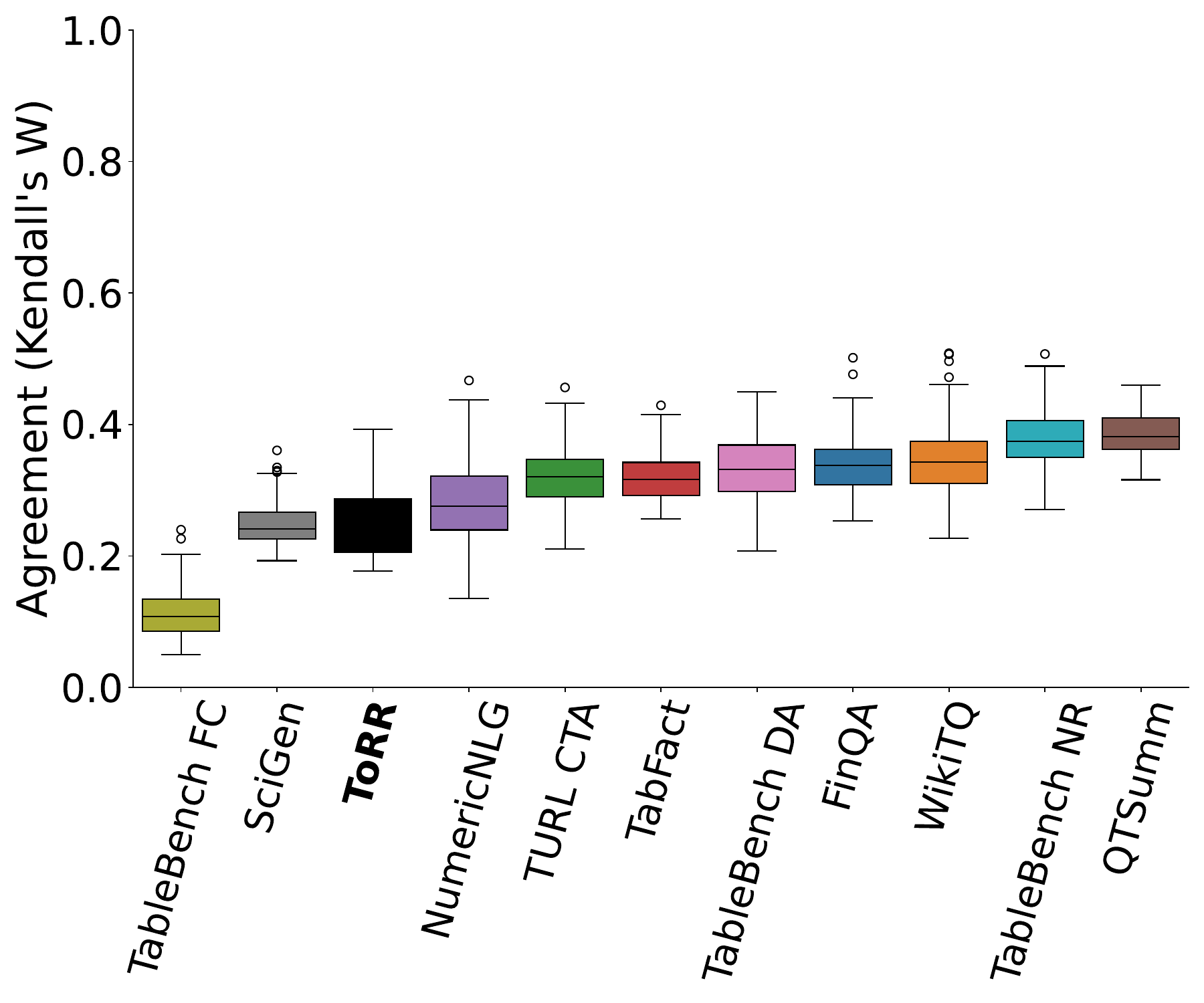}
        \caption{
        }
        \label{fig:prompt_agreement}
    \end{subfigure}
    \hfill
    \begin{subfigure}{0.42\textwidth}
        \centering
        \includegraphics[width=\textwidth]{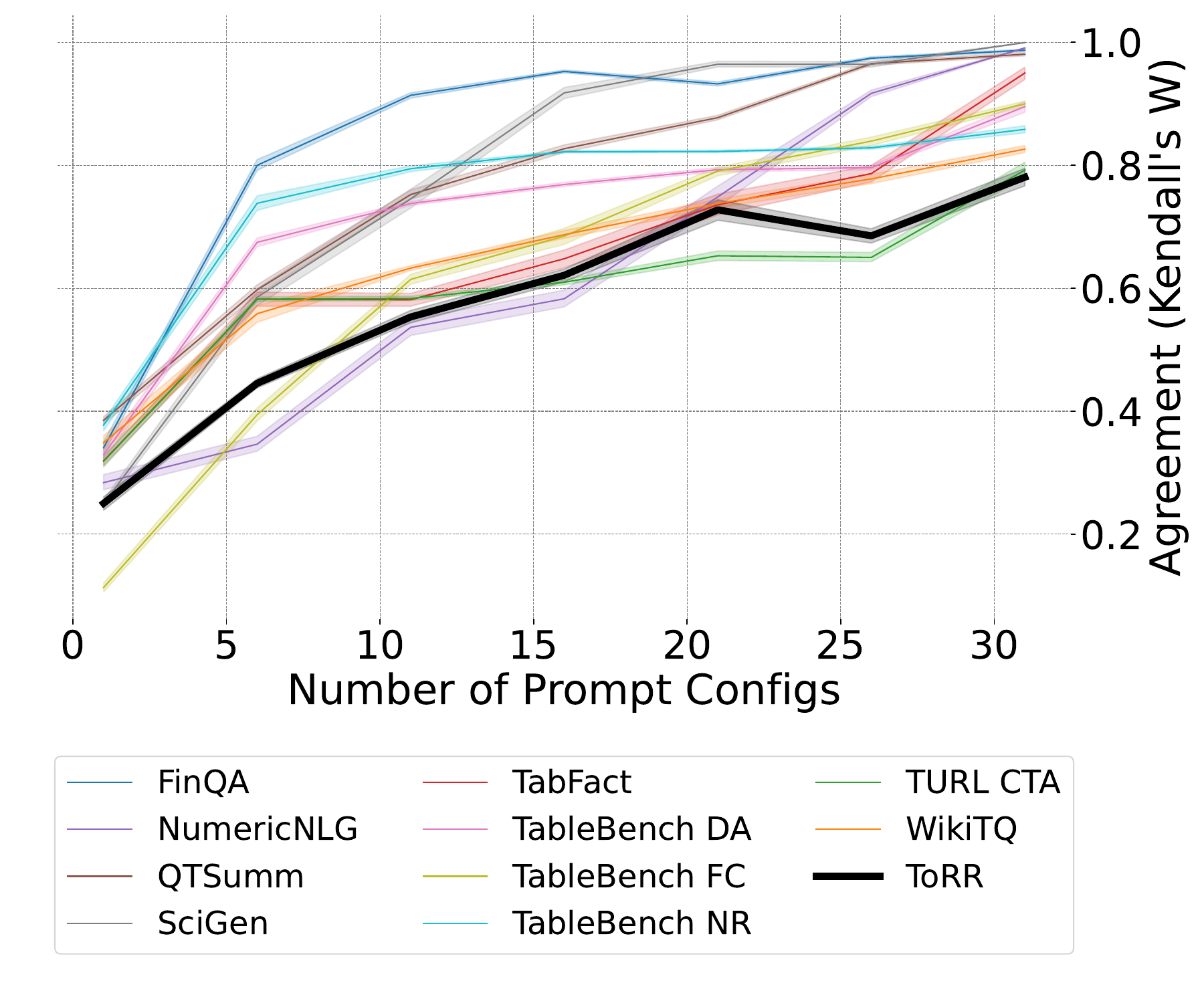}
        \caption{
        }
        \label{fig:prompt_agreement_func_prompts}
    \end{subfigure}
    \caption{Agreement between model rankings. We sample $30$ prompt configuration sets, calculate the model ranking of each set, and report the agreement (Kendall's W) between rankings. (a) \textit{Agreement between rankings based on a single prompt configuration}. Overall we see low agreement, indicating an unreliable model ranking (i.e., choosing a different prompt would lead to a different ranking). (b) \textit{Agreement between rankings based on multiple prompt configurations.} 
    Adding prompts increases the ranking consistency, with the largest gain between $\sim2$-$8$ prompts. Note that \textit{(a)} is a zoomed-in view of \textit{(b)} where the number of prompt configurations is $1$.}
    \label{fig:prompt_impact}
\end{figure*}

\paragraph{Benchmark and Dataset Separability}

We adopt the "Separability with Confidence" metric from \citet{li2024crowdsourceddatahighqualitybenchmarks}, measuring the percentage of model pairs with non-overlapping confidence intervals.\footnote{We employed bootstrapping with $1$K randomly selected seeds to sample $100$ examples from each dataset.} Separability varies across datasets (see App. Figure~\ref{fig:separability}), with \textit{WikiTQ} scoring over $71\%$ and \textit{TableBench FC} only $38\%$. Aggregated \benchmark{} achieves $79\%$, supporting the need for diverse datasets. This strengthens the need for evaluating models on tables with multiple varied datasets in order to get a reliable result. Indeed, the separability score for the aggregated \benchmark{} is significantly higher, accounting for $79\%$ by applying a similar calculation across all benchmark examples. Further interpretation is in Appendix~\ref{subsec:interpret_separability}.

\paragraph{Dataset Agreements}

To validate that \benchmark{} captures varied skills and difficulty levels, we compute model rankings per dataset and measure agreement using Kendall’s tau. Most datasets show medium to high agreement; however, Table-to-Text datasets diverge, with \textit{SciGen} and \textit{NumericNLG} aligning closely (see App. Figure~\ref{fig:sepability_agreement}).

\section{Implications for Reliable Evaluation} \label{section:prompt_impact_on_eval}

A key design choice in \benchmark{} is evaluating performance across prompt configurations (\S\ref{sec:table_forms}), which vary in table structure but preserve semantics.

While average performance is not consistently affected by prompt configuration choice (\S\ref{sec: prompt_performance}), the variability it introduces has major implications for evaluation reliability.

We use \benchmark{} to examine how prompt variation impacts the stability of model rankings, a critical aspect of benchmark reliability. Prompt configuration is one of many arbitrary design decisions that can influence evaluation outcomes \citep{perlitz2024efficientbenchmarkinglanguagemodels, reuel2024betterbenchassessingaibenchmarks}.

Following \citet{perlitz2024efficientbenchmarkinglanguagemodels}, we define reliability as ranking consistency across experimental choices. We measure this using \textbf{Kendall's W} \citep{kendall1939problem}, which quantifies agreement among rankings in the range $[0.0, 1.0]$.

\subsection{Using a Single Prompt Config is Unreliable}

Existing benchmarks usually select one prompt format (for example, serialize the table using JSON). We ask to what extent this choice influences the model ranking. Thus, we calculate model rankings based on each of our prompt configurations, and test the similarity between them.

The result is depicted in Figure~\ref{fig:prompt_agreement}. The agreement scores are generally low, demonstrating that model ranking order changes dramatically based on the choice of prompt. In other words, if a benchmark uses only a single configuration, the resulting model ranking would not be reliable.

\subsection{Multiple Prompt Configs Increase Reliability}

Figure~\ref{fig:prompt_agreement_func_prompts} depicts the effect of evaluating with multiple prompts. Unsurprisingly, we see that basing the model ranking on more prompts increases the agreement the resulting rankings have with each other. 
The plot also shows that even a relatively small number of prompt configurations can make a large difference and contribute to a more reliable model ranking. For example, increasing the number of prompts from $1$ to $10$ increases Kendall's W score by more than $0.35$ on average.

Another observation from Figure~\ref{fig:prompt_agreement_func_prompts} is that the datasets exhibit differing patterns of the increase in agreement.
For example, the agreement for \textit{FinQA} increases from $0.35$ to $0.93$ using $11$ prompts, while for \textit{NumericNLG} it increases from $0.29$ to $0.54$, suggesting that the former is more robust to prompt configurations than the latter. 
The full \benchmark{} benchmark (black line in Fig.~\ref{fig:prompt_agreement_func_prompts}) is roughly a lower bound on the prompt robustness of the model ranking across datasets.
Moreover, there appears to be a positive correlation between the task complexity suggested by our reasoning analysis (Table~\ref{tab:datasets_final_colored}) and the robustness of the datasets: FinQA, TableBench DA, and TableBench NR generally rank at the top; WikiTQ, TabFact, and QTSumm fall in the mid-range; while NumericNLG and TURL tend to show lower robustness compared to the others.


\subsection{Prompt Configs can Substitute Examples}

Figure~\ref{fig:prompt_agreement_func_prompts} shows that more prompts can help increase reliability.
A simple explanation for this is that \textit{adding more data points} helps capture model behavior. 
A common approach for this would be to add more test instances; however, increasing the size of labeled data is not always feasible.

\begin{figure}[]
  \centering
\includegraphics[width=.4\textwidth]{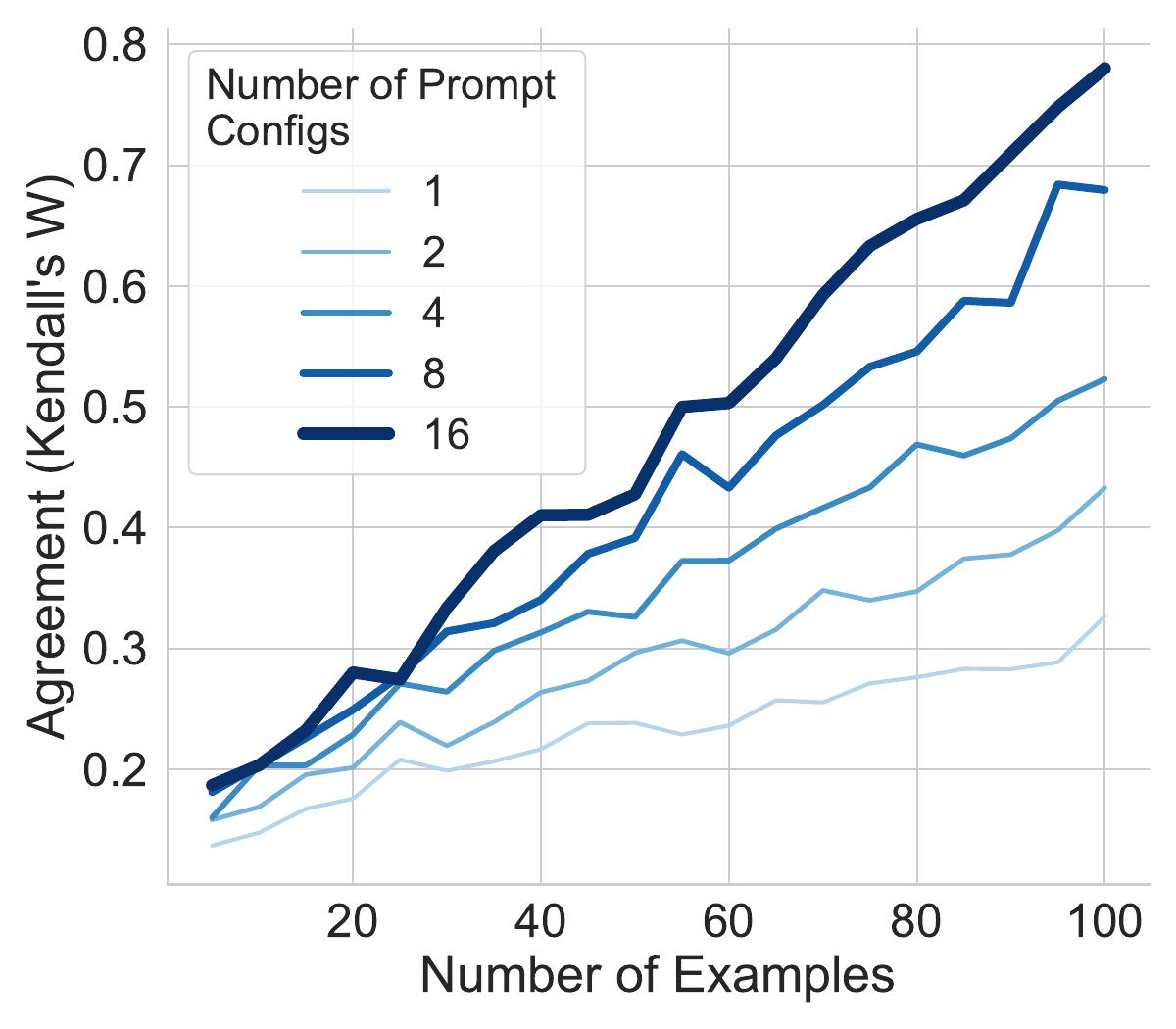}
\caption{
The improvement in model ranking consistency using different numbers of examples and prompt configurations. 
We calculate Kendall's W agreement over $30$ sets of model rankings, where each model ranking is a result of randomly selected examples and prompt configurations.
For example, for $2$ prompt configurations and $50$ examples, we repeat the following process $30$ times: randomly sample $2$ prompt configurations and $50$ examples, then calculate sample model ranking. 
}
\label{fig:comp_ex_prompts}
\end{figure}

Thus, we also look at the relation between adding examples and adding prompts.
Figure~\ref{fig:comp_ex_prompts} depicts the ranking agreement as a function of the number of test examples, averaged over each dataset.
It is evident that both the number of examples and the number of prompts consistently increase reliability. Strikingly, adding prompt variations can match the effect of doubling the test set; for instance, $50$ examples with $2$ prompts yield similar reliability to $100$ examples with one.

To conclude, evaluating multiple prompt configurations adds a valuable dimension to model assessment. While demonstrated on table formats, this approach may generalize to other tasks~\cite{liang2023holisticevaluationlanguagemodels, alzahrani2024benchmarkstargetsrevealingsensitivity, mizrahi2024state}, improving reliability and helping mitigate small test set limitations.

\section{Related Work}

Several recent works systematically evaluate LLMs on tabular data tasks, primarily focusing on question-answering type tasks.
TableBench~\citep{wu2024tablebenchcomprehensivecomplexbenchmark} 
tests LLMs over
four major categories of QA tasks, namely fact-checking, numerical reasoning, data analysis, and visualization.
DataBench~\citep{grijalba2024question}
examines
the reasoning capabilities of LLMs in a tabular context.
TQA-Bench~\citep{qiu2024tqa}, a multi-table 
benchmark, evaluates 
complex question answering 
over relational data. TabIS \cite{pang2024uncovering} evaluates the table information-seeking capabilities of LLMs. \citet{zhao2023investigating} investigate Table-to-text capabilities 
in several real-world information-seeking scenarios.
Compared to \benchmark{}, these benchmarks are limited in the tasks they cover and the model capabilities they reflect.
A detailed comparison to other table benchmarks is provided in App. Table~\ref{app:table_compare_table_benchmarks}.

More recently, there has been an increased focus on also analyzing the robustness of LLMs across table formats and perturbations.
\citet{singha2023tabular} explore the impact of table representation formats and noise operations on self-supervised table structure understanding tasks. Specifically, they consider a set of simple fact-finding and transformation tasks on tables to analyze how GPT-3 model performance varies.
Similarly, \citet{sui2024table} analyzes the capabilities of GPT-3.5 and GPT-4 in understanding tables by designing a specific set of table structure understanding tasks using structured data from various public datasets. 
Several recent works 
\citep{bhandari2024robustness, zhao2023robut, liu2023rethinking} explore structural variance and adversarial perturbations on tables, and 
their impact
on Table Question Answering 
performance.
However, existing efforts are often restricted to simple synthetic table understanding tasks or a narrow set of table QA datasets and target models, thus overlooking broader challenges posed by complex table reasoning tasks.

\section{Discussion}
In this work, we introduce \benchmark{}, the first comprehensive benchmark for table reasoning and robustness.
\benchmark{} provides a crucial resource for model developers and users seeking a more realistic and nuanced understanding of how LLMs perform in real-world tabular data scenarios.

Our results over a variety of state-of-the-art LLMs reveal a consistent pattern of relatively low performance on table reasoning tasks. 
Furthermore, and perhaps more strikingly, we demonstrate that models exhibit extreme sensitivity to seemingly minor variations in table formatting. As we show, this sensitivity does not reflect LLM preference for particular table formats; rather, the response to format variations reflects a more general phenomenon of LLM sensitivity to prompts.

This finding resonates with a growing body of recent research demonstrating the brittleness of LLMs to variations in input formatting, even in seemingly non-structured aspects of textual inputs~\citep{alzahrani2024benchmarkstargetsrevealingsensitivity}. 
These studies, while focusing on different input modalities, converge on a similar conclusion: LLMs' performance can be surprisingly brittle and inconsistent when faced with even minor input variations.
This brittleness presents a real issue for reliably evaluating LLM performance~\citep{mizrahi2024state}.

\benchmark{} directly addresses these reliability concerns by 
systematically incorporating multiple prompt configurations into the evaluation protocol.
Thus, \benchmark{}
offers a significant step towards mitigating the evaluation reliability issues, and
provides a more reliable assessment of LLM capabilities.
Moreover, utilizing multiple prompt configurations can help address the limitations of smaller test sets.

To conclude, \benchmark{} serves as a crucial benchmark for orienting future advancements in LLM table reasoning. At the same time, it sets an example for robust evaluation methodologies that are more reflective of real-world performance. We encourage the AI community to adopt this benchmark and refine the practices it introduces when developing new benchmarks.

\section{Limitations} 

\paragraph{Selected Datasets}
\benchmark{} includes datasets in which input tables are directly embedded within the prompt and can be accurately parsed into standard table formats. However, this setup does not encompass all real-world scenarios, such as cases where models must extract or search for table data independently, or handle non-standard formats like hierarchical tables.

\paragraph{Dataset Metrics} 
\benchmark{} relies on datasets and evaluation metrics created by external sources, which may introduce biases or inconsistencies that do not fully align with the intended evaluation goals.


\paragraph{Perturbation Design}
We selected four perturbations that primarily alter the order in which table content is presented in the prompt. While no semantic changes were introduced, a small number of cases may have been unintentionally affected, potentially leading to incorrect gold answers. However, limited human evaluation and additional analysis (see Appendix~\ref{app:analysis_sec:perturbs}) suggest that such instances are rare and do not significantly impact the overall results.

\section*{Acknowledgments}

We thank Ella Rabinovich for her valuable contributions in shaping the metrics used in this work. Her insights and feedback significantly improved the design of \benchmark{} and strengthened the analysis presented in this paper.

We also thank Sam Ackerman for his contributions to the statistical aspects of this work. His analysis and input helped strengthen the results of the benchmark.


\bibliography{custom}

\appendix

\appendix

\clearpage
\section{\benchmark{} Benchmark} \label{appendix:benchmark_details}

In this section, we provide more details about the benchmark and the decisions made as part of its development.

\subsection{Overview of Datasets in \benchmark{}}

\begin{enumerate}

    \item \textbf{FinQA} \citep{chen2022finqadatasetnumericalreasoning} – An expert-annotated question answering (QA) dataset designed to tackle numerical reasoning in real-world financial data. FinQA contains questions that require models to perform complex operations, such as multi-step calculations and logical reasoning over financial reports containing both text and table. \textit{License: Creative Commons Attribution 4.0 International (CC BY 4.0)}.

    \item \textbf{TableBench} \citep{wu2024tablebenchcomprehensivecomplexbenchmark} – A dataset designed to evaluate a model's table question answering capabilities across various tasks. It includes 18 distinct sub-tasks grouped into four major categories: Fact Verification (FV), Numerical Reasoning (NR), Data Analysis (DA), and visualizations. \textit{License: Apache License 2.0}.
        
    \item \textbf{WikiTableQuestions} \citep{pasupat-liang-2015-compositional} – A dataset designed for question answering over tables sourced from Wikipedia. It often requires reasoning over table data, including operations like aggregation, comparisons, and filtering, to derive accurate answers. \textit{License: Creative Commons Attribution 4.0 International (CC BY 4.0)}.
    
    \item \textbf{TabFact} \citep{chen2020tabfactlargescaledatasettablebased} – A large dataset focused on fact verification using tables, containing Wikipedia tables paired with human-annotated statements. The task is to determine whether a given statement is supported, refuted, or unverifiable based on the information in the table often requiring logical and numerical reasoning over table data. \textit{License: Creative Commons Attribution 4.0 International (CC BY 4.0)}.

    \item \textbf{QTSumm} \citep{zhao2023qtsummqueryfocusedsummarizationtabular} - A summarization dataset focused on query-based summarization, where summaries are generated based on specific user queries to retrieve relevant information from a table. \textit{License: MIT License}.

    \item \textbf{Scigen} 
    \citep{moosavi2021scigen} - 
    A dataset designed for reasoning-aware data-to-text generation, featuring tables from scientific articles along with their corresponding descriptions. \textit{License: Creative Commons Attribution-NonCommercial-ShareAlike 4.0 International License (CC BY-NC-SA 4.0)}.

    \item \textbf{NumericNLG}
    \citep{suadaa-etal-2021-towards} - 
    A dataset for table-to-text generation that pairs tables with their corresponding descriptions from scientific papers, with a focus on numerical-reasoning texts. \textit{License: Creative Commons 4.0 Attribution-ShareAlike (CC BY-SA 4.0).}.

    \item \textbf{TURL (Table Understanding through Representation Learning)} \citep{deng2020turltableunderstandingrepresentation} – The TURL Column Type Annotation (CTA) dataset, derived from Wikipedia tables, supports semantic type assignment to table columns from a given list of Freebase types. It tests models' ability to understand the meaning of table columns in context. \textit{License: Creative Commons Attribution-NonCommercial-NoDerivatives 4.0 International License (CC BY-NC-ND 4.0)}.
            
\end{enumerate}

\subsection{Other Table Dataset Types} \label{app_sec: other_table_datasets}

While we selected datasets that can be used directly in the prompt to measure the model’s ability to handle tables in the cleanest way possible, there also exist table datasets that require tool use. Our impression is that such datasets do involve the skills analyzed in this work, but they additionally depend on capabilities like tool-use formats and agentic behaviors such as search or code execution. We intentionally excluded these elements to keep the benchmark focused and interpretable, specifically targeting the model’s core tabular understanding. Moreover, many models cannot operate in such settings, and even when they can, most researchers would find the setup challenging, as evaluation frameworks are typically not designed for tool use. This design choice enables us to isolate reasoning abilities without the confounding influence of other skills. Thus, while broader coverage would allow averaging over yet another dataset, it would come at the cost of usability.

\subsection{Evaluation metrics} 

We evaluated each of the datasets using the same metrics as reported in the original papers, with some exceptions: (1) For \textbf{WikiTableQuestions}, while the original paper focuses on exact-match for scoring the prediction, we assessed models using F1 score\footnote{We tokenized both the reference and predicted tokens, where true positives are determined by the intersection of these token sets, false positives are the tokens present in the reference set but absent from the predicted set, and false negatives are the tokens in the predicted set that are missing from the reference set.}, to provide a better normalization of model performance. (2) In the case of \textbf{FinQA}, we calculated both execution accuracy and program accuracy, but used the latter as the main metric since execution accuracy tends to overestimate the performance. (3) As for \textbf{Scigen}, since the base paper states that none of the evaluated metrics align with human judgment, we adopted the primary metric of NumericNLG, as they correspond to the exact same task.


\begin{table*}[h]
    \centering
    \begin{tabular}{|l|p{10cm}|}

    \hline
    \textbf{Format} & \textbf{Example Representation} \\ \hline
    \textbf{HTML} & 
    \texttt{<table><thead>
    \newline<tr><th>Name</th><th>Age</th><th>Sex</th></tr>
    \newline</thead><tbody>
        \newline<tr><td>Sophia</td><td>26</td><td>F</td></tr>
        \newline<tr><td>Aarav</td><td>34</td><td>M</td></tr>
        \newline<tr><td>Oliver</td><td>30</td><td>M</td></tr>
    \newline</tbody>
    </table>
    } \\ \hline

    \textbf{CSV} & 
    \texttt{Name, Age, Sex\newline
Sophia, 26, F\newline
Aarav, 34, M\newline
Oliver, 30, M} \\ \hline

    \textbf{JSON} & 
    \texttt{\{``0": \{``Name": ``Sophia", ``Age": ``26", ``Sex": ``F"\}, ``1": \{``Name": ``Aarav", ``Age": ``34", ``Sex": ``M"\}, ``2": \{``Name": ``Oliver", ``Age": ``30", ``Sex": ``M"\}\}} \\ \hline

    \textbf{Markdown} & 
    \texttt{|Name|Age|Sex|\newline
|---|---|---|\newline
|Sophia|26|F|\newline
|Aarav|34|M|\newline
|Oliver|30|M|} \\ \hline

    \textbf{Indexed Row Major} & 
    \texttt{col : Name | Age | Sex row 1 : Sophia | 26 | F row 2 : Aarav | 34 | M row 3 : Oliver | 30 | M} \\ \hline

    \textbf{DataFrame} & 
    \texttt{pd.DataFrame(\{``Name": [``Sophia", ``Aarav", ``Oliver"], ``Age": [26, 34, 30], ``Sex": [``F", ``M", ``M"]\}, index=[0, 1, 2])} \\ \hline

    \textbf{Concatenation} & 
    \texttt{Name Age Sex Sophia 26 F Aarav 34 M Oliver 30 M} \\ \hline
        
    \end{tabular}
    \caption{Example representations of serialization formats used in \benchmark{}.}
    \label{tab:serialization_formats}
\end{table*}

\begin{table*}
    \centering
    \begin{tabular}{|l|p{6cm}|}

    \hline
    \textbf{Perturbation} & \textbf{Example Transformation} \\ \hline
    \textbf{Row Swapping} & \texttt{Name, Age, Sex\newline
Aarav, 34, M\newline
Oliver, 30, M\newline
Sophia, 26, F} \\ \hline
    \textbf{Column Swapping} & \texttt{Age, Name, Sex\newline
26, Sophia, F\newline
34, Aarav, M\newline
30, Oliver, M} \\ \hline
    \textbf{Transpose} & \texttt{, 0, 1, 2\newline
Name, Sophia, Aarav, Oliver\newline
Age, 26, 34, 30\newline
Sex, F, M, M} \\ \hline
    \textbf{Add Empty Rows} & \texttt{Name, Age, Sex\newline
, ,\newline
Sophia, 26, F\newline
, ,\newline
Aarav, 34, M\newline
Oliver, 30, M} \\ \hline
        
    \end{tabular}
    \caption{Structural perturbations with CSV serializer format.}
    \label{tab:perturbations_example}
\end{table*}

\subsection{Dataset Properties Evaluation}
To conduct a manual skill analysis across datasets, we sampled $15$ examples from each dataset and engaged three research scientists as annotators. Each annotator independently assessed the skills required to solve each example. A skill was considered "required" if it appeared consistently across all $15$ examples, "not required" if it was absent in all examples, and "partially required" if it was present in at least $7$ examples. Final labels were determined based on consensus, with agreement from at least two annotators deemed sufficient.

\subsection{Prompt Configurations} \label{appendix:table_forms_details}

\subsubsection{Serializations}

As tables can be presented in various formats during pre-training, models may have biases in how they interpret them as demonstrated by \cite{sui2024table}. Several standard methods exist for converting structured data into text formats that can be embedded in a prompt. We select seven of the most commonly used serializations: \textit{HTML, CSV, JSON, Markdown, Indexed Row Major, Data Frame and Concatenation}. An example representation of each serialization format is provided in Table \ref{tab:serialization_formats}.

\subsubsection{Structural Perturbations} \label{subsec: structural_perturb}

These perturbations present a slightly different structure of the table from the original one, without any change to the content or relations between the table cells. Since the information is preserved but presented in a different form, there should be no significant change in the model performance ideally. The perturbations considered are as follows:
\begin{itemize}
    
    \item \textbf{Row Swapping}: The process of exchanging the positions of rows within a table.
    
    \item \textbf{Column Swapping}: The act of changing the positions of columns in a table.

    \item \textbf{Transpose}: The action of switching the rows and columns of a standard table, transforming rows into columns and vice versa. 

    \item \textbf{Add Empty Rows}: An addition of some empty rows to the table.
\end{itemize}

Refer to Table \ref{tab:perturbations_example} for an example of these structural perturbations with CSV as a base serializer.

\subsection{Data Contamination}\label{app:sec_contamination}

A general concern in benchmarking is the issue of data contamination, where the benchmarked models have memorized some of the data in the benchmark. Our approach in \benchmark{} is likely to mitigate some of the effects of contamination due to the use of $35$ distinct prompt variations per example. Specifically, we expect the effects of memorization to be tied to the exact phrasing of an input example; thus, using an unseen prompt format the model is less likely to utilize memorized data. Hence, over a diverse set of prompts we do not expect a model to succeed based solely on memorized data. In other words, our derived performance metrics implicitly penalize shallow memorization, offering a more reliable evaluation of true generalization capabilities (e.g., if a model succeeds on one specific format of an example due to memorization, it will get a performance score of $2.8\%$ on that example).


Moreover, the diversity of datasets in terms of domains, creators, and other attributes further contributes to reducing the likelihood of contamination.

Nevertheless, as is the case for other benchmarks, we cannot rule out some effects of data contamination on the results.

\subsection{Challenges}

Building a robust benchmark for table reasoning like \benchmark{} presented several challenges, particularly related to dataset quality and evaluation consistency.

\paragraph{Memorized Data.} 
Many existing table datasets are derived from the same sources, particularly Wikipedia tables, which large language models have likely seen during pre-training. As a result, these datasets may be less challenging, as models can rely on memorization rather than genuine reasoning.

\paragraph{Evaluation Rigor.}
The evaluation metrics used in prior work do not always accurately reflect model performance. Several studies indicate that automatic metrics often fail to align with human evaluations, leading to misleading conclusions about model capabilities. This inconsistency makes it difficult to assess true reasoning ability and robustness.

\paragraph{Unreliable Data.} 
Some datasets rely on automatically aggregated structured data, which can introduce inconsistencies and errors. For instance, a dataset with high potential for benchmarking included web tables, but the extracted table HTML was broken at times, making it unreliable for structured reasoning tasks as such issues can compromise the integrity of the benchmark.

\subsection{Computation Cost} \label{app:computation_cost}
The full ToRR benchmark uses between approximately $110$ million and $140$ million tokens per model under evaluation. Based on the pricing on the model inference platforms used for this paper (OpenAI, Anthropic, Google Vertex AI and Together AI) as of May 2025, the total cost of model inference API credits for all evaluations was approximately $\$2{,}400$.

\clearpage




\section{Usage}\label{appendix:usage_options}

\benchmark{} can be easily run using \texttt{unitxt}. After installing the package, users can reproduce our results with the following code snippet:

\begin{figure}[h]
    \centering
    \includegraphics[width=0.7\textwidth]{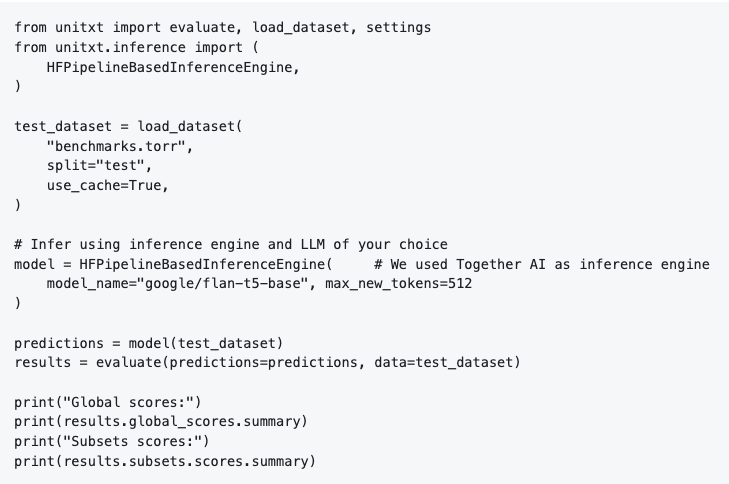}
    \caption{Example code for running \benchmark{} using \texttt{unitxt}.}
    \label{fig:placeholder}
\end{figure}

\newpage
\null\newpage
\section{Full Names}

\subsection{Model Names}\label{appendix:model_names}

\begin{table}[h]
\centering
\begin{tabular}{ll}
\hline
\textbf{Full Model Name} & \textbf{Short Name} \\
\hline
qwen2-72b-instruct & qwen2-72b-i \\
mixtral-8x22b-instruct-v0.1 & mixtral-8x22b-i \\
mixtral-8x7b-instruct-v0.1 & mixtral-8x7b-i \\
mistral-7b-instruct-v0.3 & mistral-7b-i \\
llama-3-1-70b-instruct & llama-3-1-70b-i \\
llama-3.1-70b-instruct & llama-3-1-70b-i \\
llama-3.1-405b-instruct & llama-3-1-405b-i \\
llama-3.1-8b-instruct & llama-3-1-8b-i \\
gpt-4o-mini-2024-07-18 & gpt-4o-mini \\
gpt-4o-2024-11-20 & gpt-4o \\
gemini-1.5-flash-002 & gemini-1.5-flash \\
gemini-1.5-pro-002 & gemini-1.5-pro \\
claude-3-5-sonnet-20241022 & claude-3-5-sonnet \\
claude-3-5-haiku-20241022 & claude-3-5-haiku \\
\hline
\end{tabular}
\caption{Mapping of model names to short names.}
\label{tab:model_names}
\end{table}

\subsection{Dataset Names}\label{appendix:dataset_names}

\begin{table}[h]
\centering
\begin{tabular}{ll}
\hline
\textbf{Full Dataset Name} & \textbf{Short Name} \\
\hline
TURL - Column Type Annotation & TURL CTA \\
TableBench - Data Analysis & TableBench DA \\
TableBench - Numerical Reasoning & TableBench NR \\
TableBench - Fact Verification & TableBench FC \\
\hline
\end{tabular}
\caption{Mapping of dataset names to short names.}
\label{tab:dataset_names}
\end{table}

\clearpage
\section{Statistical testing of models} \label{appendix:statistical_testing}

Here we present results from a statistical analysis of model performance.  We analyze the results of the 14 models listed in Table~\ref{table:perform} on the ten datasets listed in Table~\ref{tab:datasets_final_colored}. The analysis follows the procedure in \citet{ackerman2025statisticalmultimetricevaluationvisualization}.  

We define a single run configuration for an LLM as the unique combination of values of the serializer, augmenter, and example index.  For each dataset, the average metric score of the model on each configuration is calculated.
This yields $3{,}500$ unique configurations for each dataset, except for TableBench FC, which had $2{,}765$. Following the analysis setup in \citet{ackerman2025statisticalmultimetricevaluationvisualization}, a new dataset is formed for each source dataset in Table~\ref{tab:datasets_final_colored}, where each observation in the dataset corresponds to a particular configuration, and its (single) score is the average metric value for the configuration.  Thus, for instance, we form the dataset $D_1$ of size $3{,}500$ (unique configurations) for source dataset FinQA.  

Since for each model we have results for all configurations,
we can directly compare the results for a pair of models on a given configuration via a paired (observation-level) hypothesis test.  
For each dataset indexed $j$, and pair of models $(a,b)$, we obtain a single p-value $p_{a,b,j}$ and effect size $e_{a,b,j}$, which reflect how significantly different the pair of models are across the configurations, on that dataset $j$. Effect sizes are often used as an alternative to p-values because they are less influenced by the sample size (i.e., the number of configurations) and aim to reflect whether the difference has a practical meaning.
The p-values are the nonparametric paired Wilcoxon, adjusted to control the false discovery rate with $\alpha=0.05$ \cite{benjamini2001control}.  

We find that \textit{within each dataset}, some model pairs were not statistically significantly different using the p-value method, after adjusting for multiplicity.



We also calculate \textit{aggregated} p-values ($p_{a,b}$) and effect sizes ($e_{a,b}$) across all the datasets. Following \citet{ackerman2025statisticalmultimetricevaluationvisualization}, p-values are aggregated using Wilson's harmonic mean method \citep{wilson2019harmonic} and effect sizes using the inverse variance-weighted mean \citep{turner2006calculating}.

After aggregating the results across all datasets, we find that all pairwise model comparisons are significantly different from each other according to the p-value ($p<0.05$). However, using the effect size criterion (paired Cohen's $d$), which better reflects practical differences, nearly all pairwise effects are relatively small (smaller than $0.5$). 


\clearpage

\section{Performance Analysis} \label{app:perform_analysis}

This part of the paper presents additional analyses and graphs illustrating model performance.

\subsection{Model Performance across Datasets}
\begin{figure}[h]
    \centering
    \includegraphics[width=0.7\linewidth]{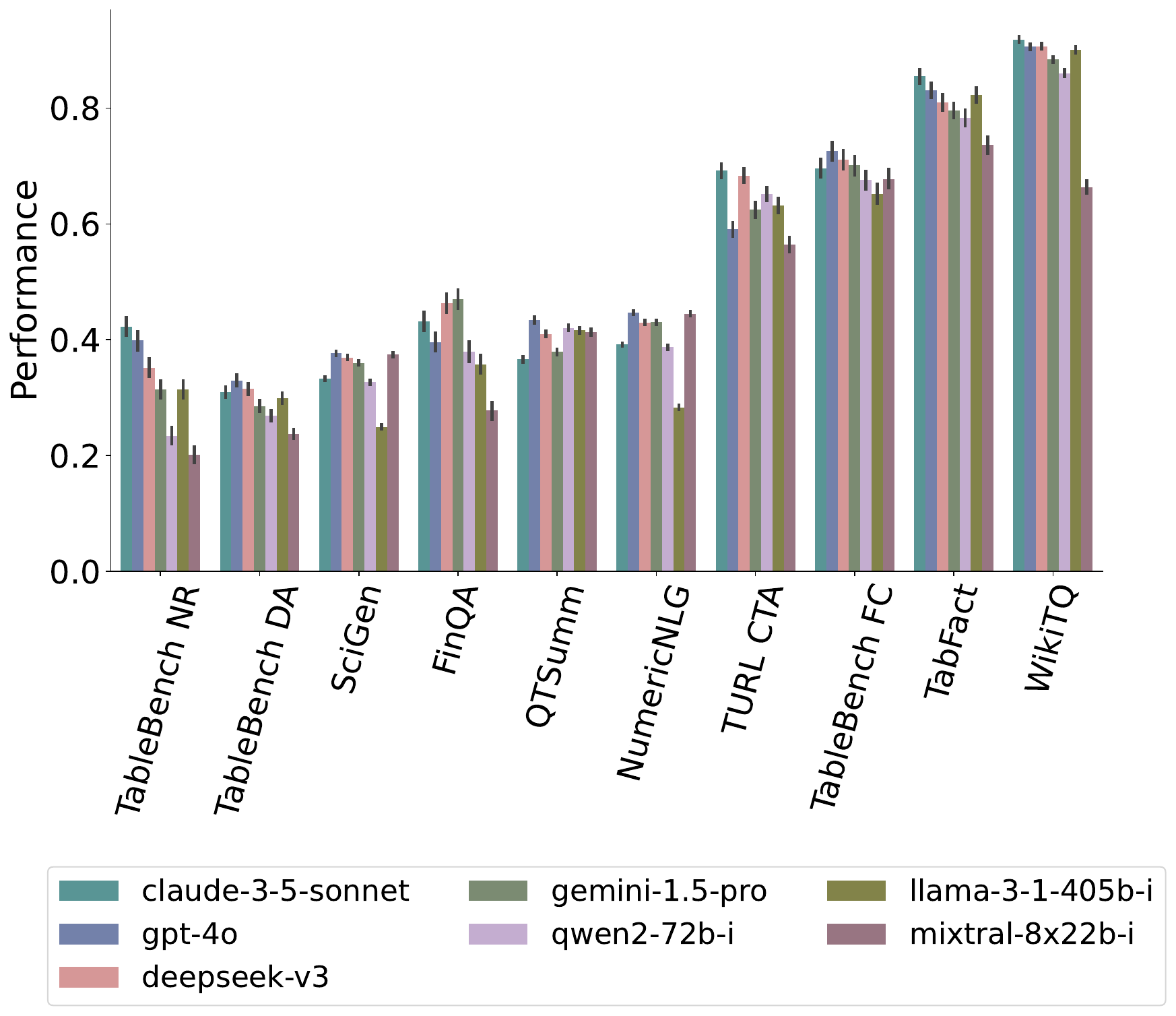}
    \caption{Performance comparison of the strongest model from each family.}
    \label{fig:model_performance}
\end{figure}

Figure~\ref{fig:model_performance} compares the performance of all models across each dataset in the benchmark. In general, performance tends to be lower on more challenging datasets- FinQA, TableBench Numerical Reasoning, TableBench Data Analysis as well as Table2Text ones. On the other hand, performance scores are relatively higher on question answering and fact-checking datasets including WikiTQ, TabFact and others.

\subsection{Open vs. Closed models}

\begin{figure}[]
  \centering
\includegraphics[width=0.45\textwidth]{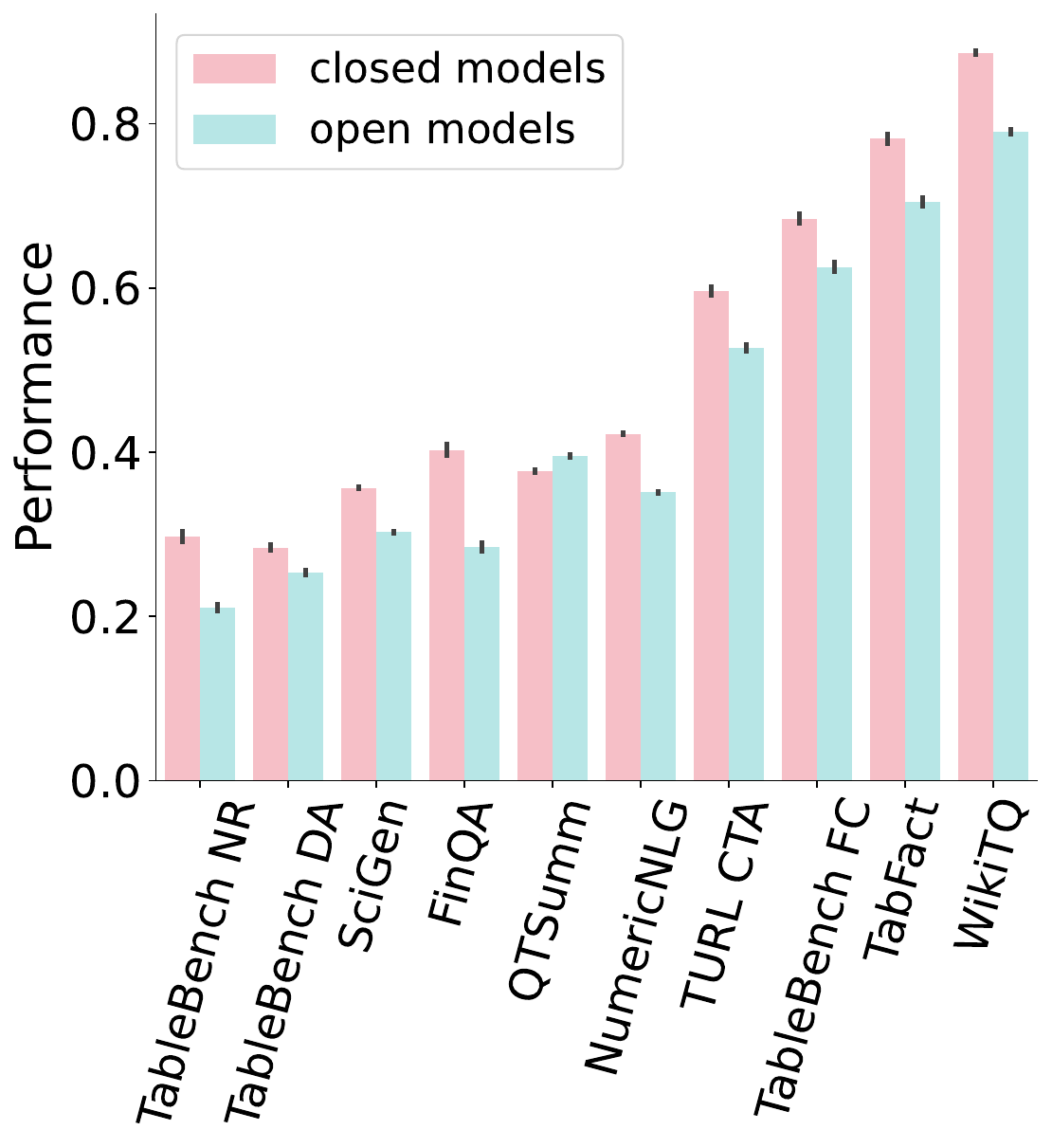}
\caption{Comparison of open weight and closed models.}
\label{fig:open_closed}
\end{figure}

Figure~\ref{fig:open_closed} presents the performance comparison between open-weight and closed-weight models across all the benchmark datasets. As can be seen, closed-weight proprietary models considerably outperform open models across most benchmark datasets.


\clearpage
\section{Robustness Analysis} \label{app:robust_analysis}

In this section, we present additional analyses of the key components of prompt configurations: serializers and structural perturbations.

We provide both performance scores and win rate graphs. All scores are calculated in a manner similar to that described in \ref{subsec:metrics}. Win rate calculations were performed at the example level: for each model's score on an example, we counted the number of times it was higher than the others, and then normalized it by the total number of wins over this example. For example, given the following scores of $7$ serializers across an example: $[1, 1, 1, 0, 0, 0, 0]$, the counting vector will be $[4, 4, 4, 0, 0, 0, 0]$, and the normalized win rate scores will be $[1/3, 1/3, 1/3, 0, 0, 0, 0]$.

\subsection{Serializers Impact} \label{app:analysis_sec:serializers}

Here, we provide more in-depth analysis of trends concerning serializers. We approach this in steps: starting with the lowest level of aggregation for each model-dataset pair, we then move to model-oriented and dataset-oriented aggregations, and finally conclude with an overall aggregation of the serialization results.

\begin{table*}[b]
    \resizebox{\textwidth}{!}{
    \begin{tabular}{lrrrrrrrrrr}
\toprule
 & FinQA & \makecell{Numeric-\\NLG} & QTSumm & SciGen & \makecell{TURL\\ CTA} & \makecell{Tab \\ Fact }& \makecell{TableBench \\ DA} & \makecell{TableBench \\FC} & \makecell{TableBench \\NR} & WikiTQ \\
\midrule
claude-3-5-haiku & .04 & .01 & .01 & .01 & .02 & .06 & .02 & .07 & .11 & .05 \\
claude-3-5-sonnet & .02 & .01 & .01 & .00 & .02 & .03 & .01 & .08 & .07 & .02 \\
deepseek-v3 & .04 & .02 & .01 & .00 & .02 & .06 & .02 & .03 & .04 & .04 \\
gemini-1.5-flash & .03 & .01 & .10 & .00 & .02 & .08 & .01 & .04 & .06 & .03 \\
gemini-1.5-pro & .04 & .01 & .03 & .01 & .04 & .05 & .01 & .04 & .05 & .05 \\
gpt-4o & .05 & .00 & .01 & .00 & .04 & .04 & .01 & .04 & .05 & .03 \\
gpt-4o-mini & .06 & .01 & .01 & .01 & .03 & .09 & .01 & .05 & .06 & .03 \\
llama-3-1-405b-i & .04 & .07 & .02 & .06 & .04 & .04 & .05 & .23 & .06 & .03 \\
llama-3-1-70b-i & .02 & .07 & .01 & .03 & .04 & .06 & .04 & .20 & .04 & .04 \\
llama-3-1-8b-i & .03 & .01 & .02 & .01 & .17 & .07 & .07 & .22 & .07 & .02 \\
mistral-7b-i & .06 & .04 & .02 & .04 & .03 & .08 & .01 & .14 & .04 & .04 \\
mixtral-8x22b-i & .04 & .01 & .02 & .01 & .02 & .04 & .01 & .09 & .06 & .06 \\
mixtral-8x7b-i & .05 & .03 & .03 & .02 & .16 & .06 & .02 & .11 & .03 & .09 \\
qwen2-72b-i & .06 & .07 & .01 & .05 & .02 & .03 & .01 & .06 & .03 & .02 \\
\bottomrule
\end{tabular}
    }
    \caption{The largest variations in model scores across different serializers, presented for each dataset.}
    \label{table:serializers_score_diff}
\end{table*}

\subsubsection{Model-Dataset Score Variability}
To measure the extent of variability with respect to serializers, we calculated the score for each model-dataset pair across all serializers and computed the difference between the highest and lowest scores. The result is shown in Table~\ref{table:serializers_score_diff}.
While on average the drop in score is about $0.05$, there are some exceptional cases were it can be much worst, e.g. \textit{llama-3-1-8b-i} has a drop in score of $0.22$ over \textit{TableBench FC}.

\subsubsection{Serializer Preference across Models}
\begin{figure*}
\includegraphics[width=0.75\textwidth]{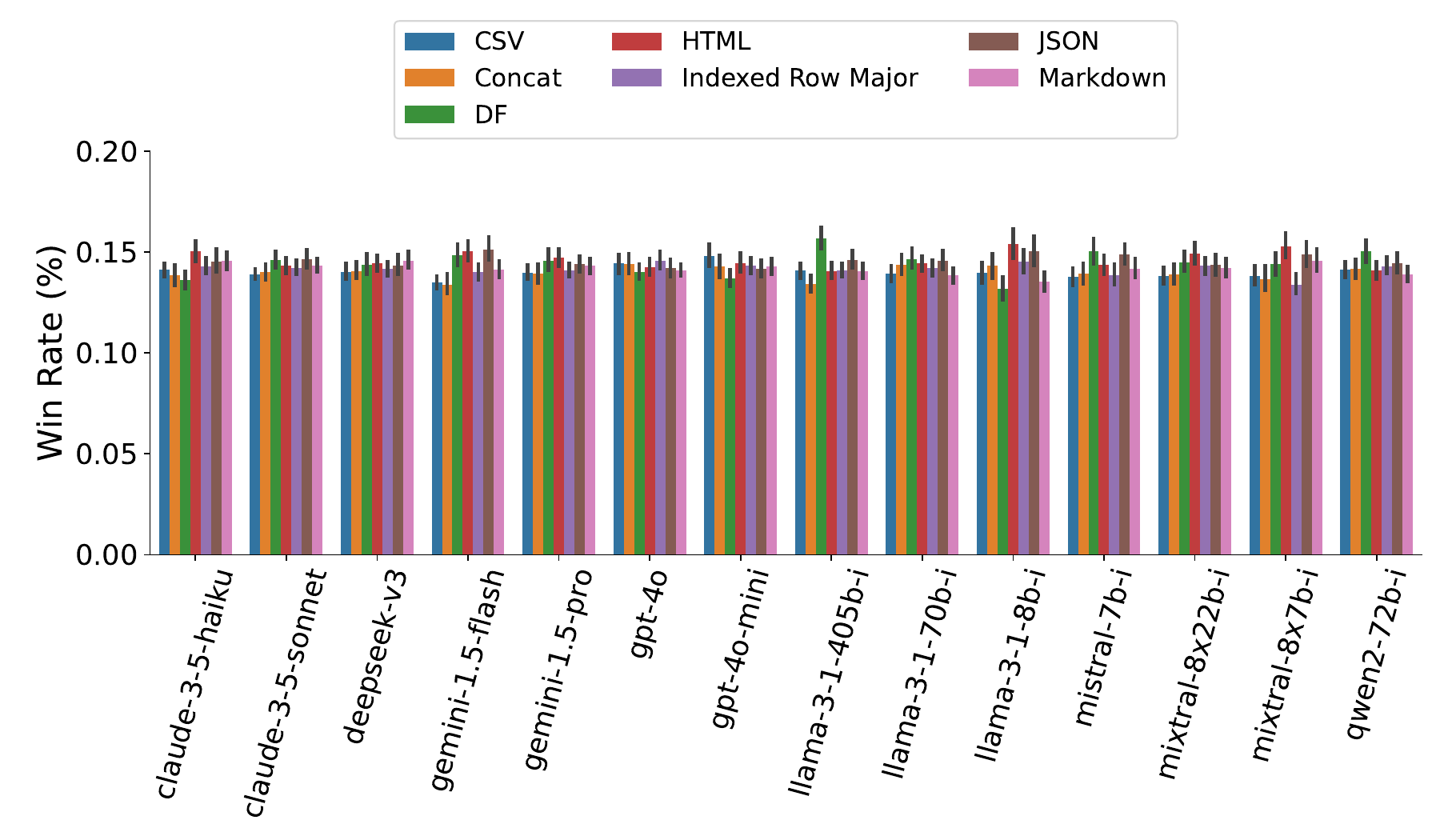}
\caption{Model-wise win-rate comparison of serializers in \benchmark{}. Some models don't show clear preferences, like \textit{gpt-4o}, while others show a preference for one or more serializers. For example, \textit{llama-3-1-405b-i} prefers DF serialization.
}
\label{fig:serializers_win_across_models}
\end{figure*}

To identify trends related to models, we aggregate the results across models, as illustrated in Figure \ref{fig:serializers_win_across_models}. This figure shows the serializer preferences of all models within our benchmark, indicating that different models exhibit varying preferences, with no single serializer consistently ranking highest across all models. 
Overall, since preferences are highly model-dependent, selecting the most suitable serializer may require case-by-case tuning.

\subsubsection{Serializer Preference across Datasets}
\begin{figure*}
\includegraphics[width=0.75\textwidth]{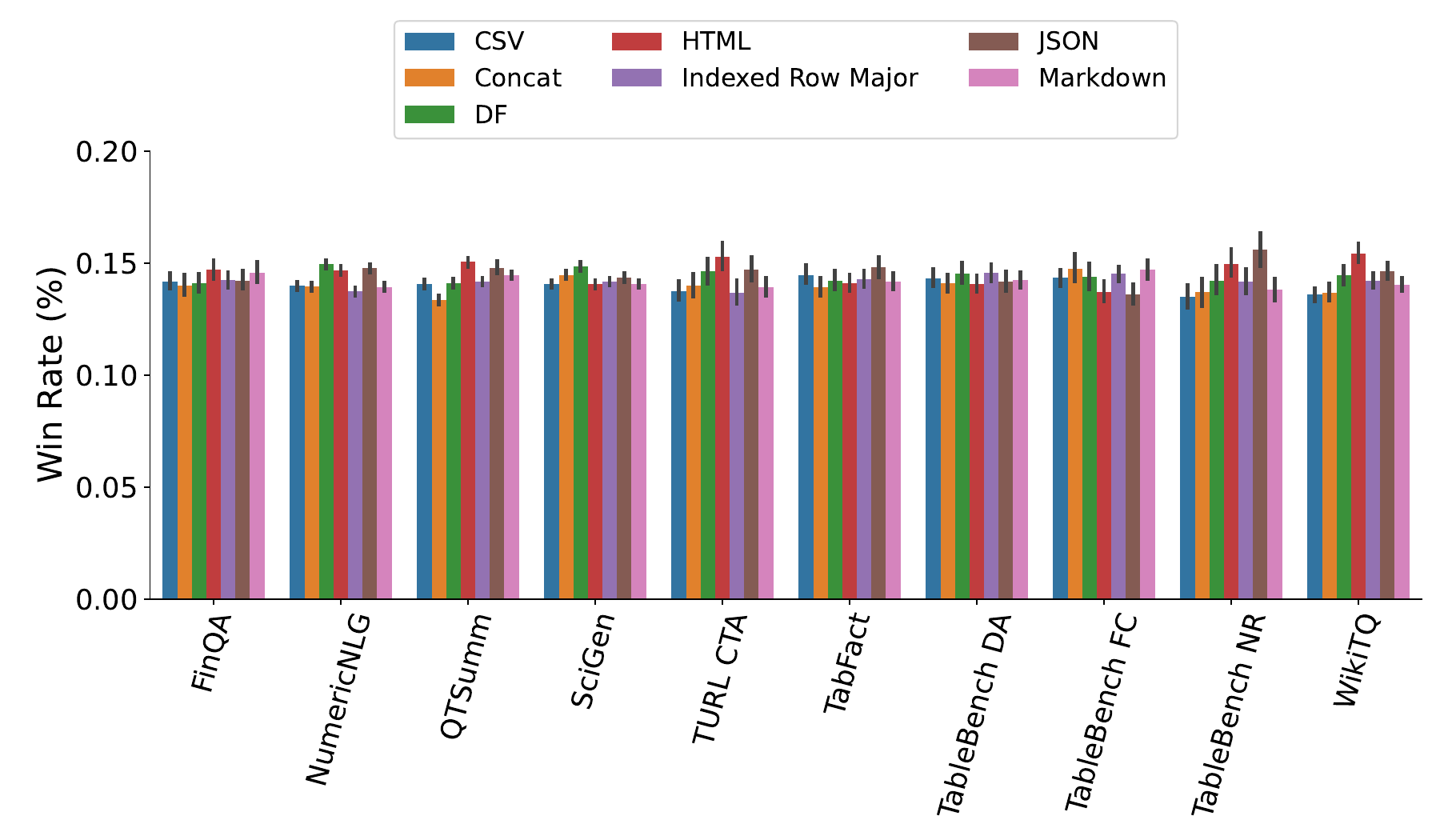}
\caption{
Dataset-wise win-rate comparison of serializers in \benchmark{}. Surprisingly, the \textit{concat} serialziation, which may render the prompt unreadable due to the lack of detailed separation, doesn't present a dramatically low performance across datasets and tasks.
}
\label{fig:serializers_win_across_datasets}
\end{figure*}

We explored signals indicating which serializers perform better for different tasks and datasets, as shown in Figure \ref{fig:serializers_win_across_datasets}.
The results show that no single serializer consistently outperforms others across all datasets. In some cases, the ranking differences among serializers is marginal, while in others, a clear performance gap emerges. This emphasizes the need for dataset-specific evaluation rather than a one-size-fits-all approach.

\subsubsection{Overall Serializer Preferences}

The results presented above show a lack of consistency in preferences. Unsurprisingly, aggregating all the results across models and datasets
demonstrates that no serializer consistently outperforms others.


\begin{figure*}[]
\centering
\includegraphics[width=0.35\textwidth]{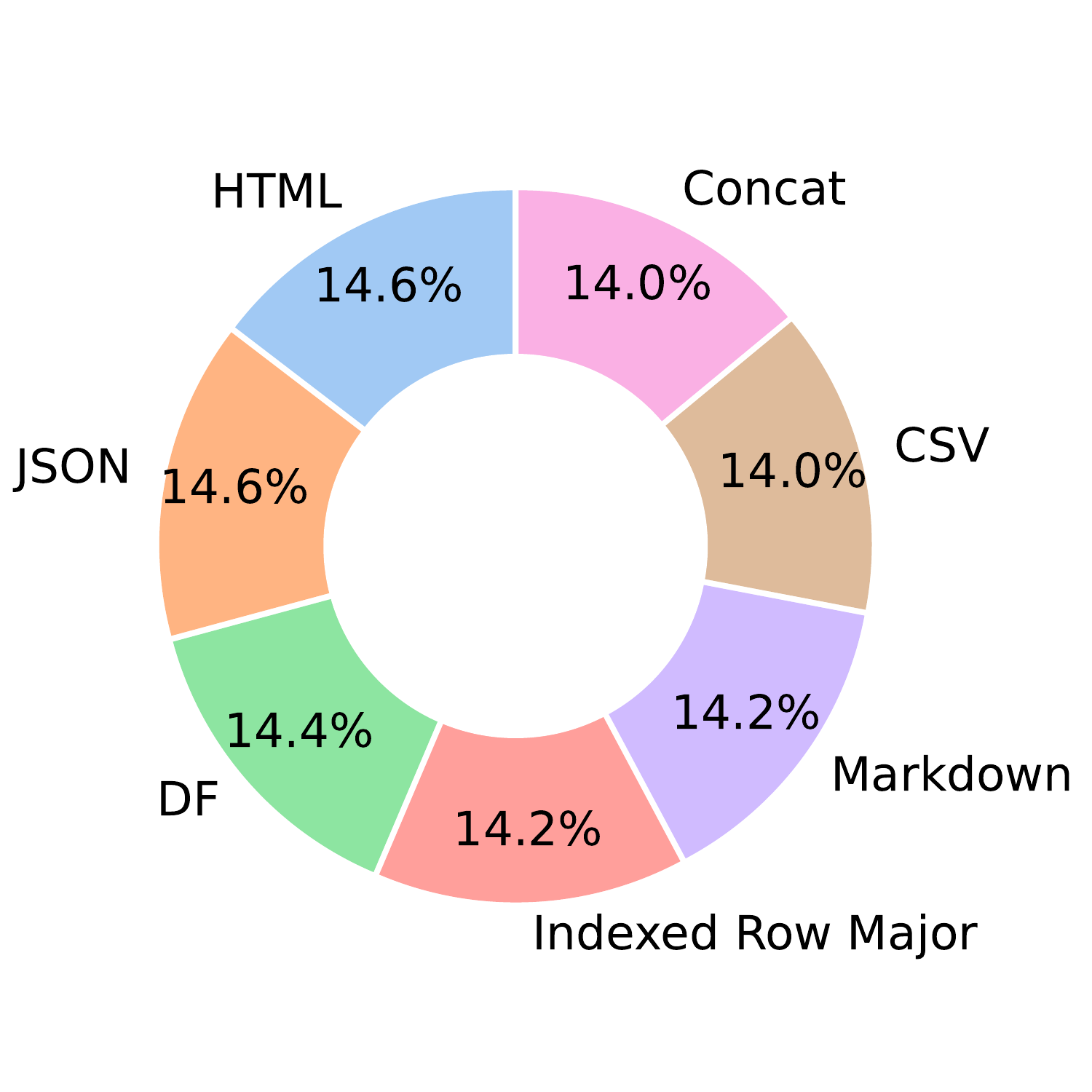}
\caption{Win rate of serializers averaged across all models and datasets.
}

\label{fig:serializers}
\end{figure*}

\clearpage
\subsection{Perturbations Impact} \label{app:analysis_sec:perturbs}

\begin{figure*}[b]
    \centering
    \includegraphics[width=\textwidth]{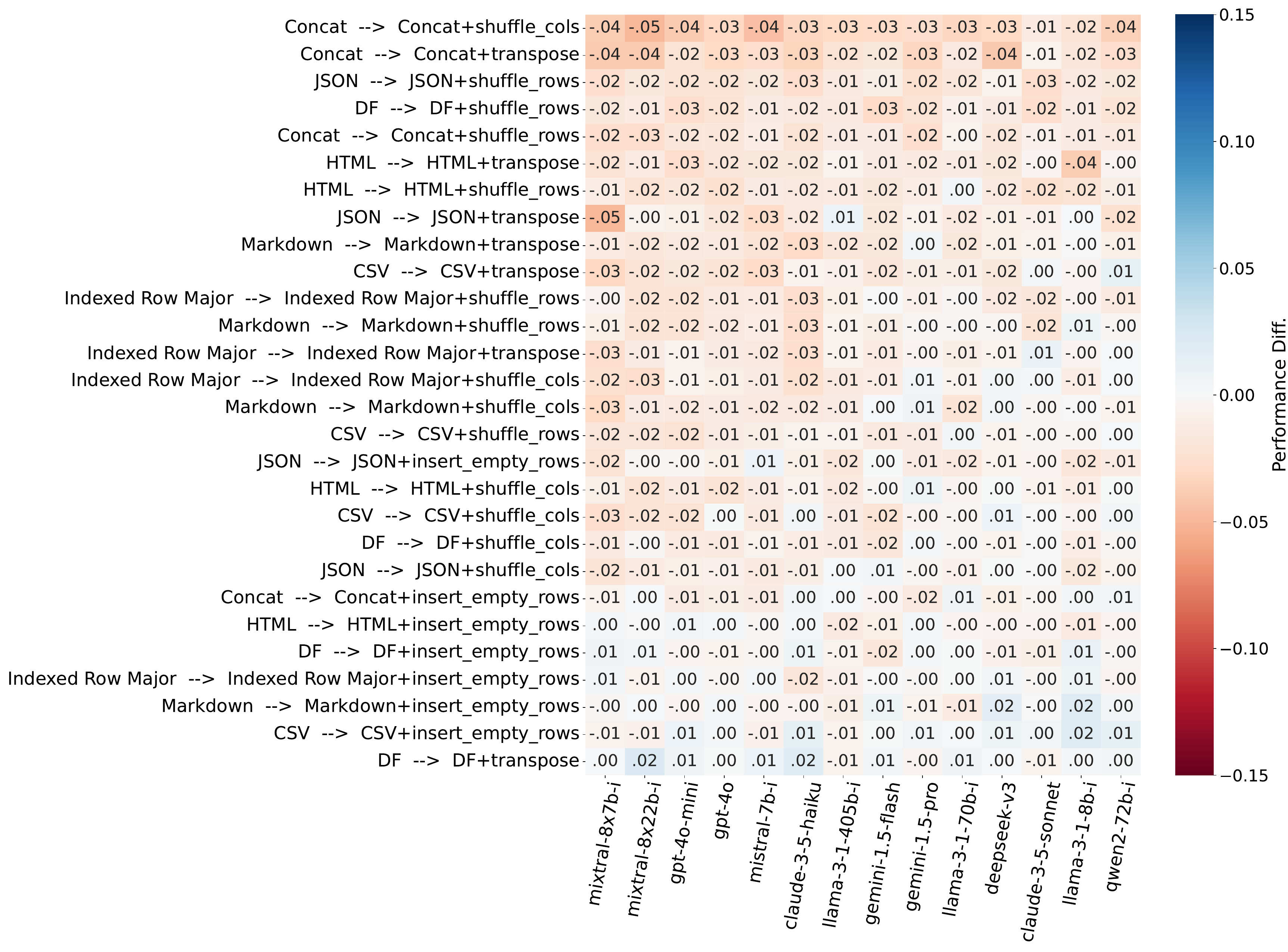}
    \caption{Differences in model score for perturbations with respect to the serializer. Overall, the effect of each perturbation seems to be very low and not consistent.}
    \label{fig:perturb_no_abs_models}
\end{figure*}

\begin{figure*}
\centering
\includegraphics[width=0.9\textwidth]{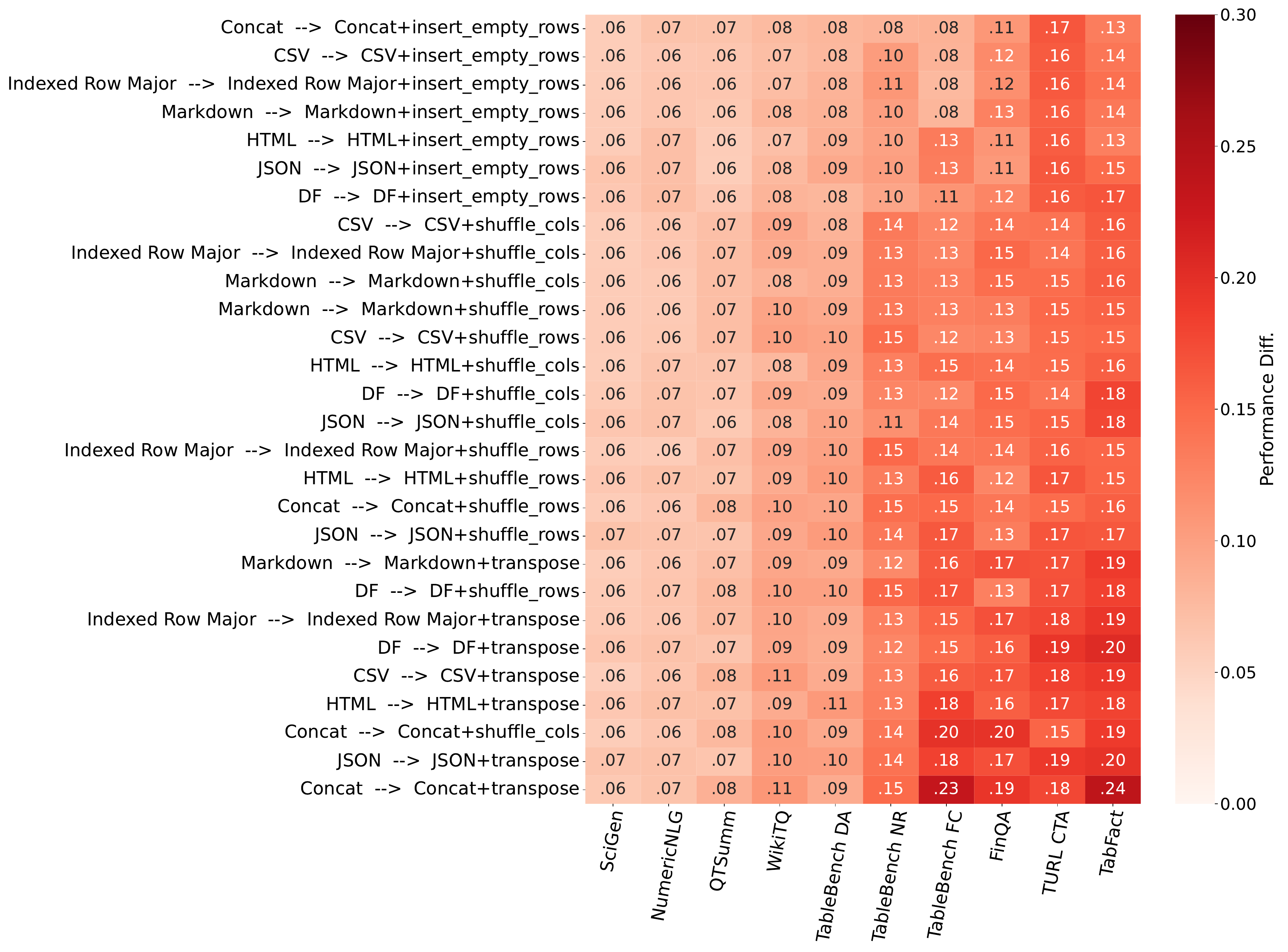}
\caption{Mean Absolute Impact of different perturbations on benchmark datasets. The signal of differences in score appears to be a part of the datasets rather than related to perturbations.
}
\label{fig:compare_perturbs_dataset}
\end{figure*}

\begin{figure*}
\centering
\includegraphics[width=0.9\textwidth]{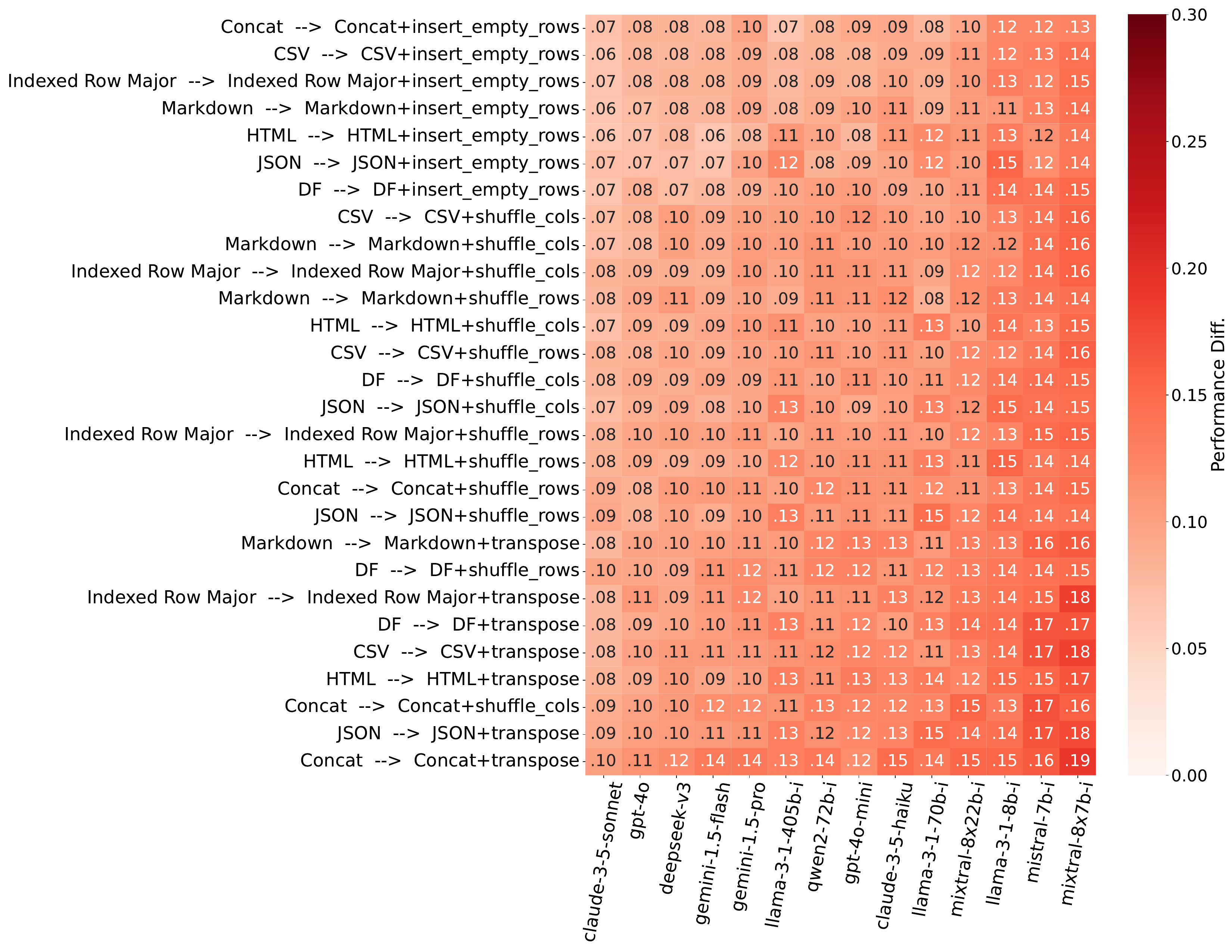}
\caption{Mean Absolute Impact on each model performance due to perturbations. The average model robustness score (presented in Table~\ref{table:perform}) appears to correlate with the model's robustness across different perturbations.}
\label{fig:compare_perturbs_model}
\end{figure*}

Here, we present a more detailed analysis of trends associated with the structural perturbations in \benchmark{}. We begin with the low-level aggregations of the differences between scores with and without perturbations, and then we aggregate the overall results.

\subsubsection{Perturbations Impact}
To gain deep insights into the impact of perturbations on performance, we analyzed the score differences introduced by each perturbation across benchmark models. 
We established a baseline using prompts with a specific serializer and no perturbations, and then measured the score deviations caused by each perturbation in comparison to this baseline.

The result, depicted in Figure~\ref{fig:perturb_no_abs_models}, reflects small yet inconsistent variability in model scores. Models exhibit an average difference of $0.03$ in overall performance.

\subsubsection{Granular Look: Absolute Perturbations Impact}

While some perturbations exhibit a strong positive or negative impact on certain examples, their effects vary inconsistently across all datasets and models. 
A simple average, as we see in Figure~\ref{fig:perturb_no_abs_models}, can be misleading as effects may cancel out. Here we report Mean Absolute Impact, which quantifies the overall effect using absolute score changes ($\Delta$). Let us denote $i \in N$ the number of examples, and $\Delta_i$ the difference between observed and expected scores of perturbation with respect to its baseline. The Mean Absolute Impact is defined as: \[ \text{Mean Absolute Impact}  = \frac{1}{N} \sum_{i=1}^N |\Delta_i| \]

Figure~\ref{fig:compare_perturbs_dataset} shows the Mean Absolute Impact on performance scores due to different perturbations for each benchmark dataset, averaged across all models. 
The impact is higher for Table QA and Fact-checking datasets and lower for Table-to-Text datasets. Similarly Figure~\ref{fig:compare_perturbs_model} displays the Mean Absolute Impact on performance scores for each model, averaged across all benchmark datasets.
The impact of perturbations is more pronounced in smaller models compared to larger models in general. Overall, it seems that both datasets and models are key factors that influence robustness.

\subsubsection{Overall Perturbations Effect}

\begin{figure}[h]
  \centering
\includegraphics[width=0.45\textwidth]{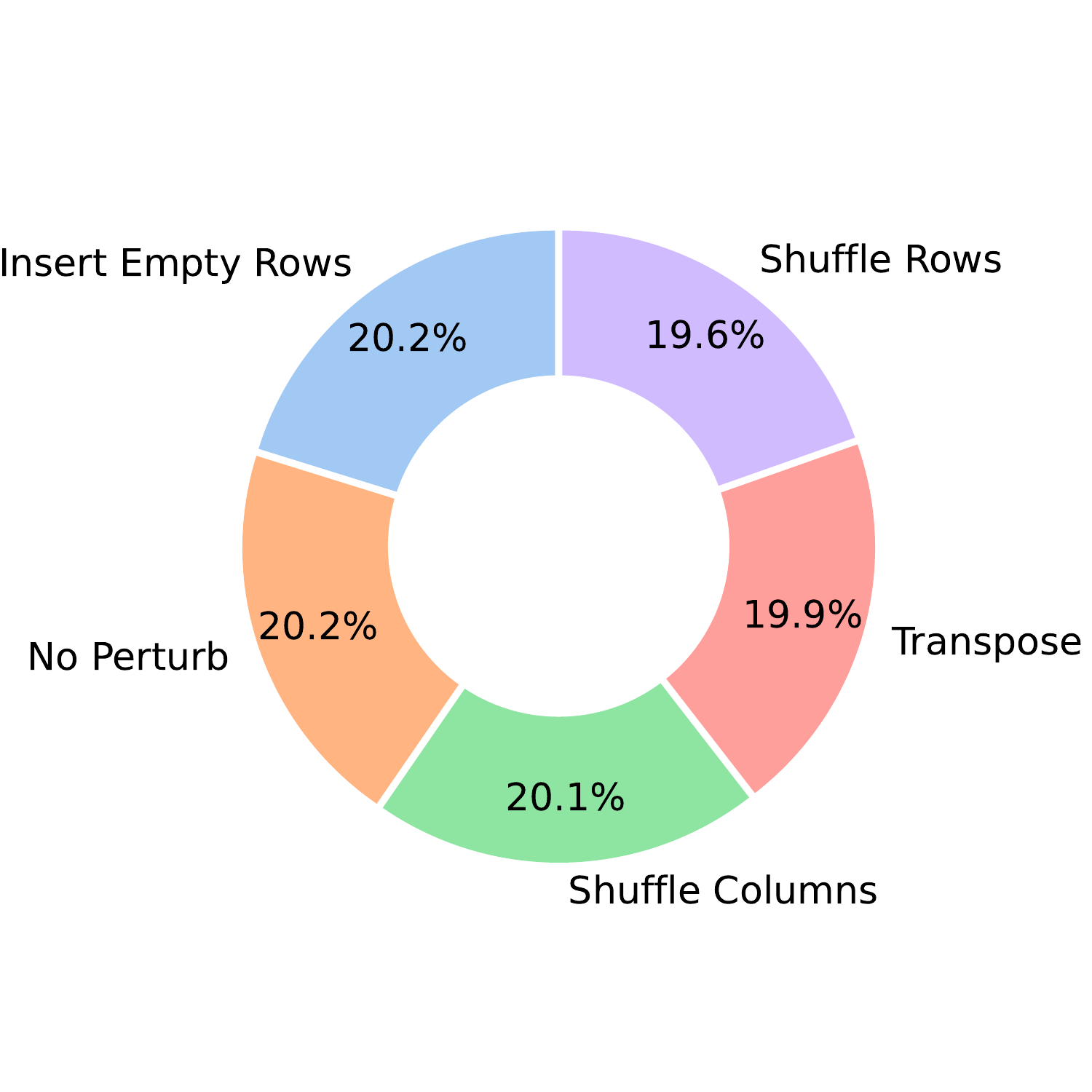}
\caption{Win rate of structural perturbations across all models and datasets of \benchmark{}.
}
\label{fig:augmenter}
\end{figure}

Figure \ref{fig:augmenter} presents the win-rate of structural perturbations, representing the percentage of times a perturbation outperforms all others aggregated across the benchmark. The win rate results are nearly identical across all perturbations, indicating that no single perturbation consistently leads to under-performance or out-performance of models when aggregated across the benchmark datasets and models.

\clearpage

\clearpage
\section{Additional Properties of \benchmark{}}

Another perspective on benchmark variability can be gained by examining it at the example level. We present two analysis that relate to example difficulty and can serve for indicating better on this variability in \benchmark{}.

\subsection{Score Distribution in \benchmark{}}

\begin{figure}[h]
     \centering
    \includegraphics[width=0.5\textwidth]{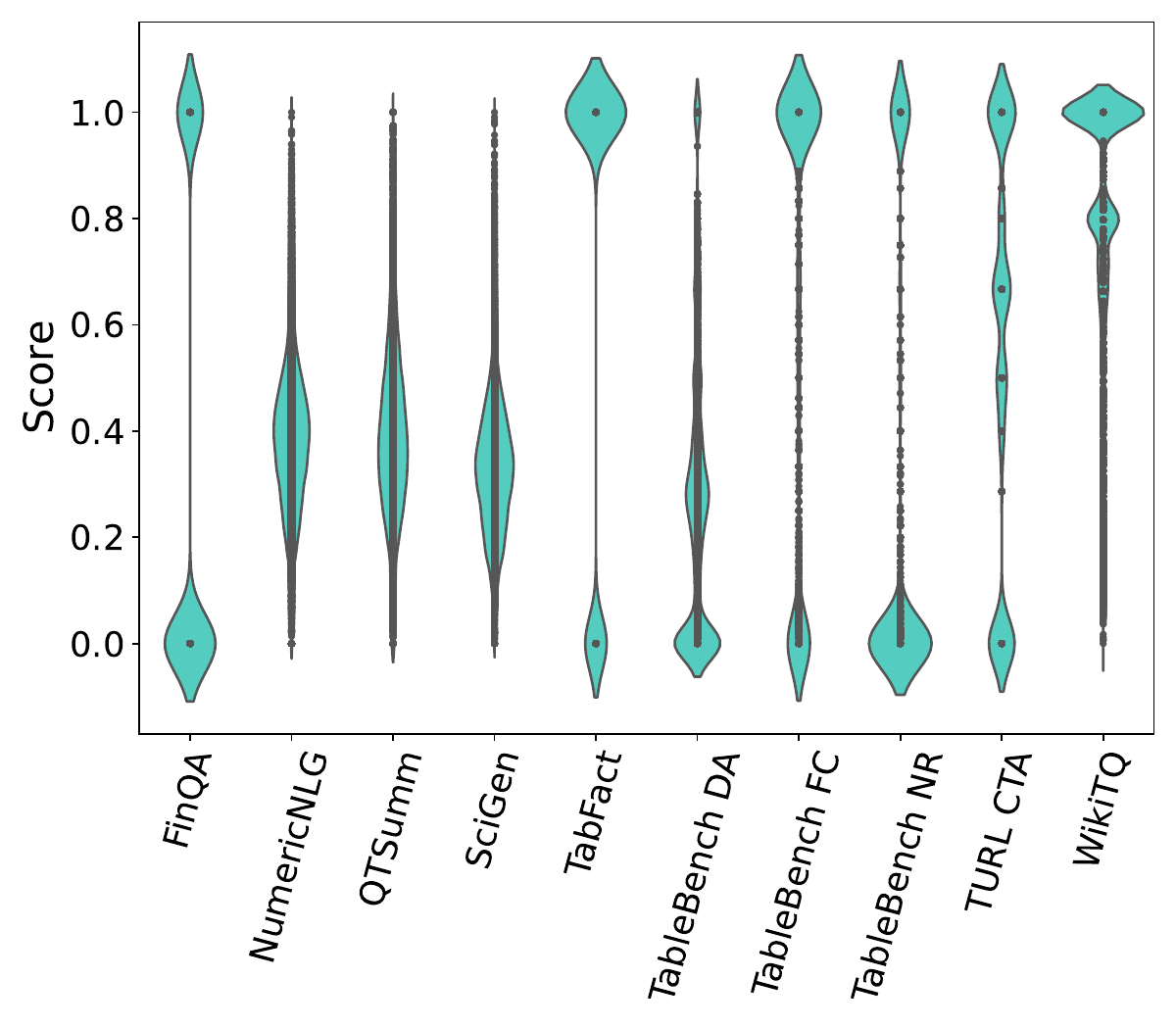}
    \caption{Score distribution across \benchmark{} datasets.
    }
    \label{fig:example_dist}
\end{figure}

To gain insight into the difficulty of each dataset, we analyze the scores achieved by the models across different prompt configurations, focusing on the score distribution. The distribution patterns show notable differences, even when datasets were evaluated with the same metric, as shown in Figure \ref{fig:example_dist}.
For example, datasets like TableBench, NumericNLG, and QTSumm, which all use \textit{ROUGE}, exhibited distinct score patterns. This suggests that the dataset itself has a strong and varied impact on model performance.

\subsection{Example Difficulty in \benchmark{}}

\begin{figure}[h]
    \centering
    \includegraphics[width=0.5\textwidth]{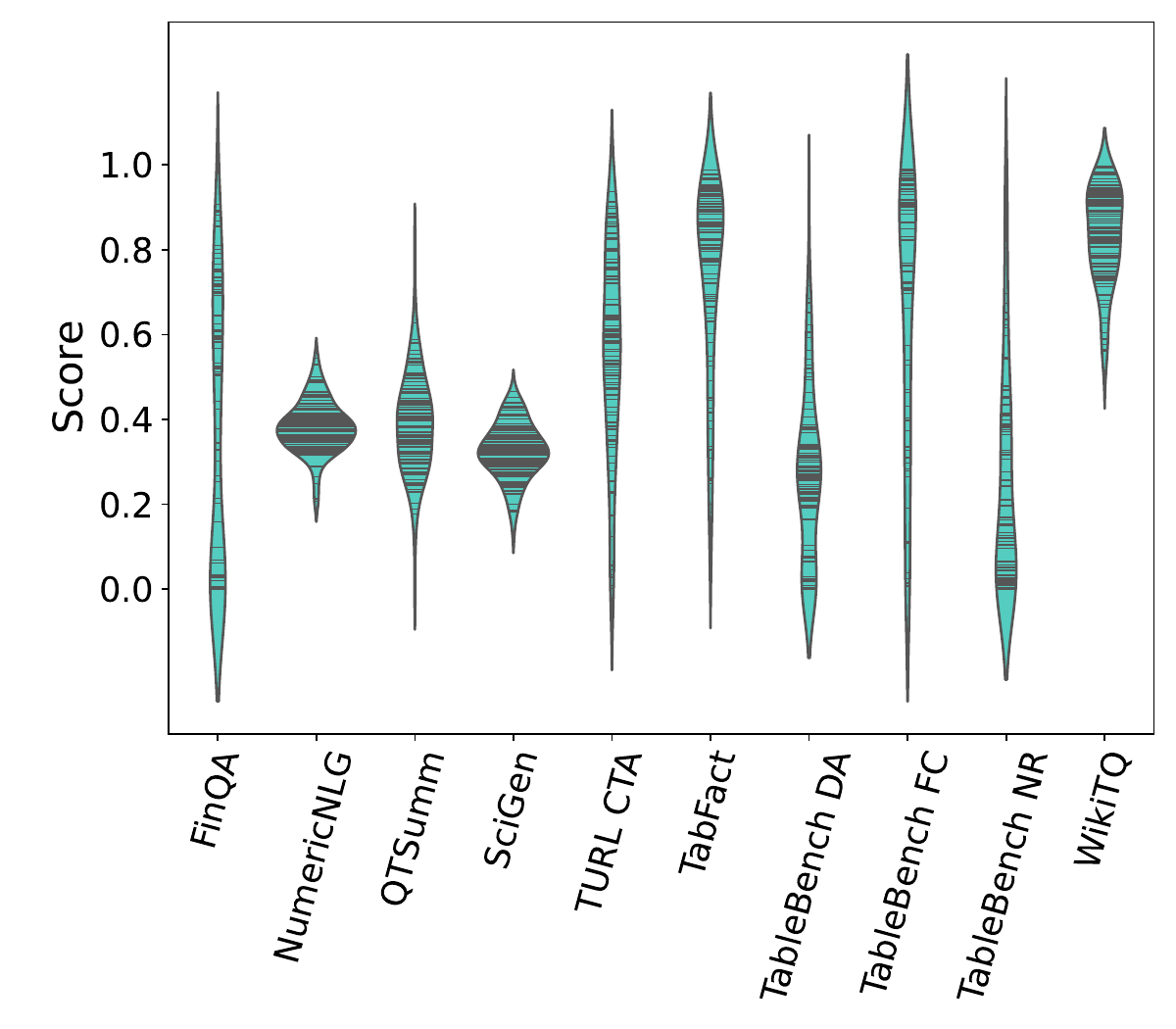}
    \caption{Example difficulty distribution for each dataset in \benchmark{}. Some datasets like \textit{FinQA} and \textit{TableBench} ones include examples with different levels of difficulty for the models, while others like Table-to-Text ones seem to include examples with a similar level.
    }
    \label{fig:example_dist_difficulty}
\end{figure}

We also analyze the difficulty of examples within our benchmark by computing the mean score for each example across all models and prompt configurations. We then examine the distribution of these aggregated mean scores across datasets, as illustrated in Figure \ref{fig:example_dist_difficulty}. The figure indicates that our benchmark encompasses both easy and challenging examples, with the majority falling within a medium difficulty range.

\subsection{An Analysis for Interpreting Dataset Separability} \label{subsec:interpret_separability}

\begin{figure*}
    \centering
    \includegraphics[width=\textwidth]{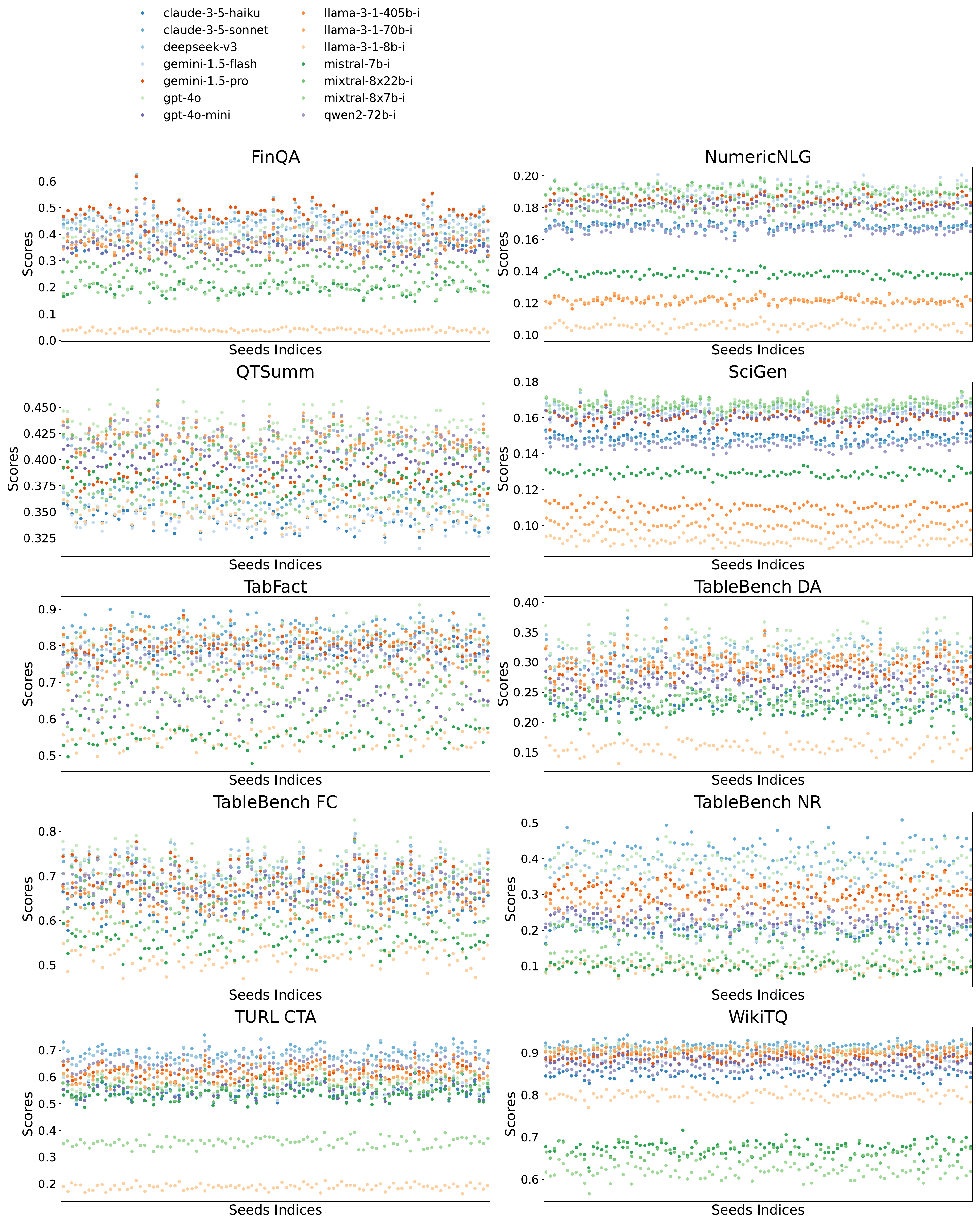}
    \caption{Each graph shows the score distribution and variability across models for a given dataset, based on bootstrapping with different seeds. This visualization supports the separability analysis, as it highlights how distinctly model scores spread within each dataset.}
    \label{fig:seperability_interpretability}
\end{figure*}

\begin{figure}[h]
  \centering
         \includegraphics[width=0.45\textwidth]{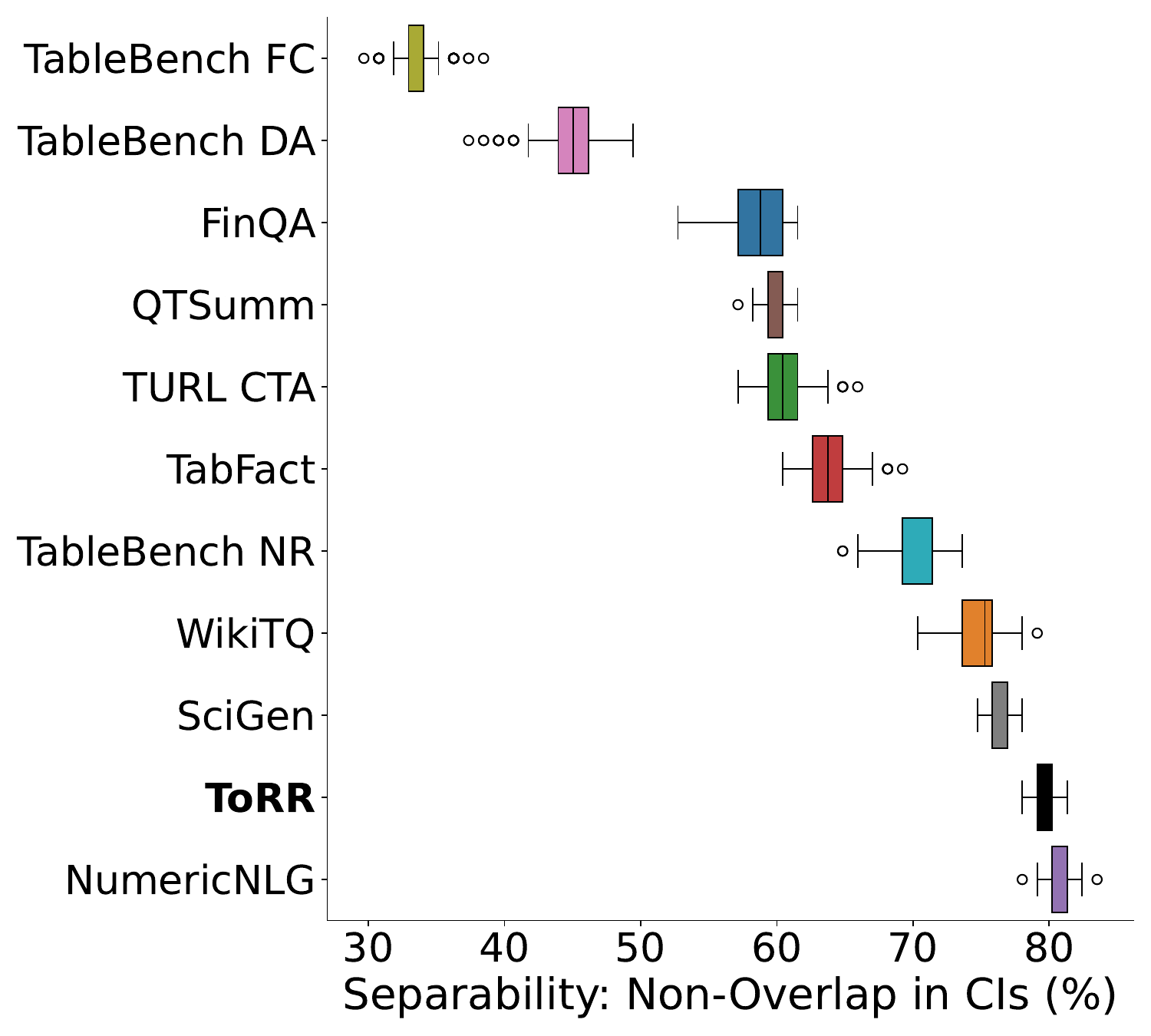}
         \caption{The separability score for each dataset in \benchmark{}. This score represents the proportion of model pairs that can be distinguished with confidence, meaning their confidence intervals (CIs; via bootstrapping over $1$K seeds) do not overlap.}
         \label{fig:separability}
\end{figure}

As can be seen in Figure~\ref{fig:separability}, some datasets exhibit higher separability values, while others do not, and this does not strongly correlate with other analyses or results we obtained. To gain intuition about why this occurs, we visualized the behavior of model scores for each dataset using $100$ different seeds through bootstrapping — following the same method we used to compute separability, and the results are shown in Figure~\ref{fig:seperability_interpretability}.

The figure suggests that this score is influenced by several factors; The overall score distribution across the dataset, the selected models and the score differences between them, and the variation in difficulty and diversity of the dataset's questions. For example, \textit{NumericNLG} and \textit{SciGen} appear more homogeneous in terms of question types, resulting in stable model scores and, consequently, higher separability.

In contrast, \textit{WikiTQ} shows a more expected trend: weaker models tend to vary more in their scores than stronger ones, and the observed high separability may be a result of the selection of models. \textit{TableBench NR}, on the other hand, seems to be challenging for all models (i.e., the score range is approximately $0.15-0.40$) but also contains a wide variety of questions, as indicated by the high variability in model scores.

\section{Related Work}

\begin{table}[h]
\centering
\renewcommand{\arraystretch}{1.3}
\begin{tabular}{|p{1.7cm}|p{2cm}|p{3cm}|p{1.2cm}|p{1.5cm}|p{1.5cm}|p{2cm}|}
\hline
\textbf{Benchmark} & \textbf{Focus Area} & \textbf{Focus Tasks} & \textbf{Task Diversity} & \textbf{Format Robustness} & \textbf{Model Types} & \textbf{Key Uniqueness} \\
\hline
ToRR (ours) & Table Reasoning and Robustness Analysis & Table2Text, Summarization, Column Annotation, QA (Fact Checking, Numerical Reasoning, Data Analysis) & High – 6 tasks, 10 datasets & Yes – Multiple formats, perturbations & General LLMs & Robustness benchmark for diverse table reasoning tasks \\
\hline
TableBench \cite{wu2024tablebenchcomprehensivecomplexbenchmark} & Table Question Answering & QA only (Fact Checking, Numerical Reasoning, Data Analysis, Visualization) & Moderate & No – Fixed formats & General LLMs & Performance ranking on QA \\
\hline
InfiAgent-DABench & Data Analysis with Agents & Data analysis questions & Narrow & No – Agent execution focus & Agents + LLMs with tools & Agent planning for data analysis \\
\hline
TableVQA-Bench & Table Visual Question Answering & Visual table QA (images) & Narrow & No – Image-based format & Vision-Language Models & Vision-language reasoning \\
\hline
DataBench \cite{grijalba2024question} & Table Question Answering & Table QA & Narrow & No – Fixed representation & General LLMs & QA performance evaluation \\
\hline
TQA-Bench \cite{qiu2024tqa} & Multi-table QA over relational data & Table QA & Narrow & Limited – 2 formats & General LLMs & QA over interconnected relational tables \\
\hline
\end{tabular}
\caption{Comparison of Table Reasoning Benchmarks.}
\label{app:table_compare_table_benchmarks}
\end{table}

\newpage
\null\newpage
\section{\benchmark{} Properties} \label{app:properties}

\begin{figure}[h]
  \centering
 \includegraphics[width=0.45\textwidth]{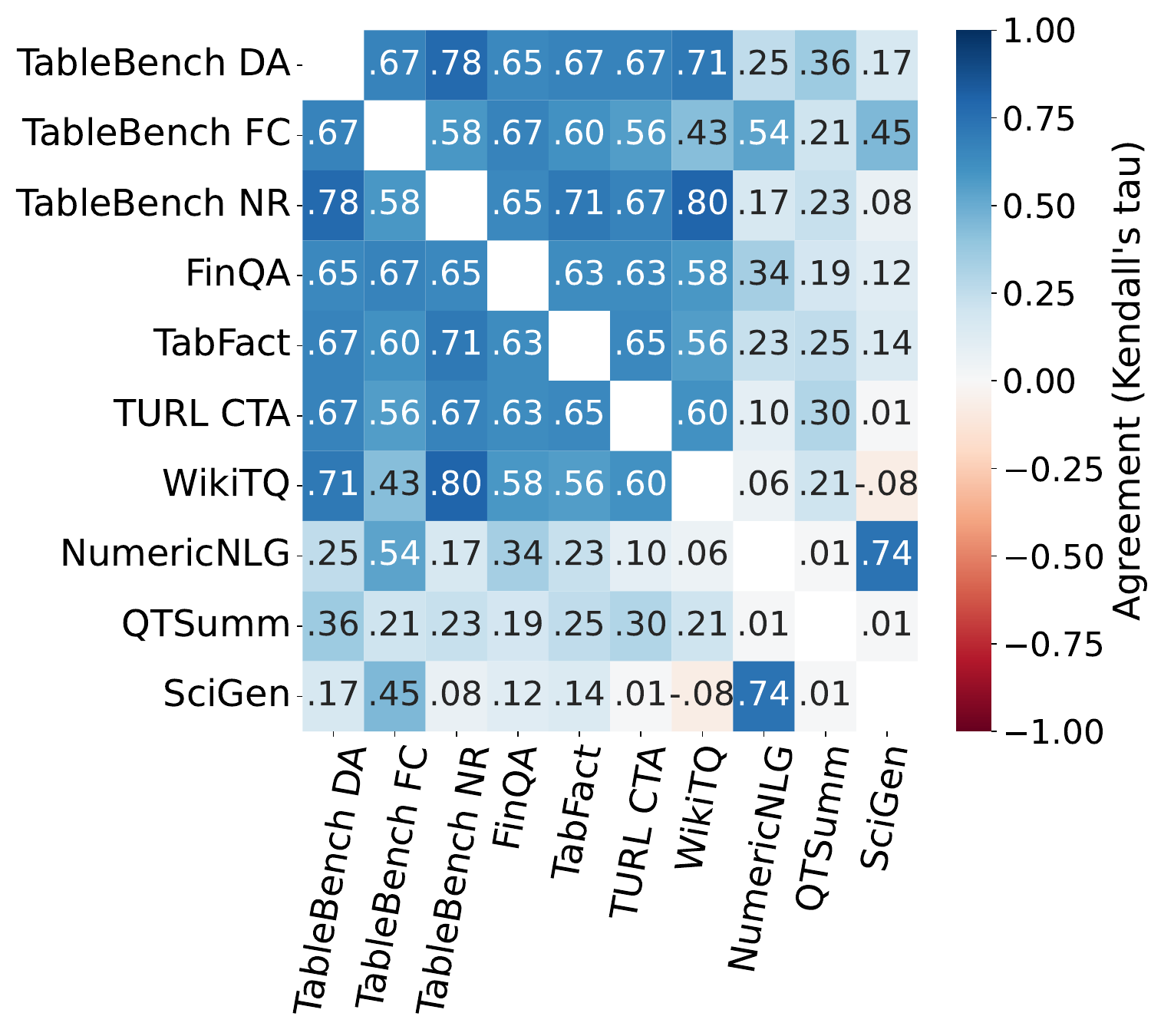}
 \caption{Model ranking agreement between the datasets in \benchmark{}.}
 \label{fig:sepability_agreement}
\end{figure}

\section{Usage in AI}
In this work, we used AI models exclusively for language-related tasks, such as rephrasing and surface-level linguistic transformations. 

\newpage
\null\newpage
\section{Prompt Examples}

\begin{figure*}[h]
    \centering
    \includegraphics[width=\textwidth]{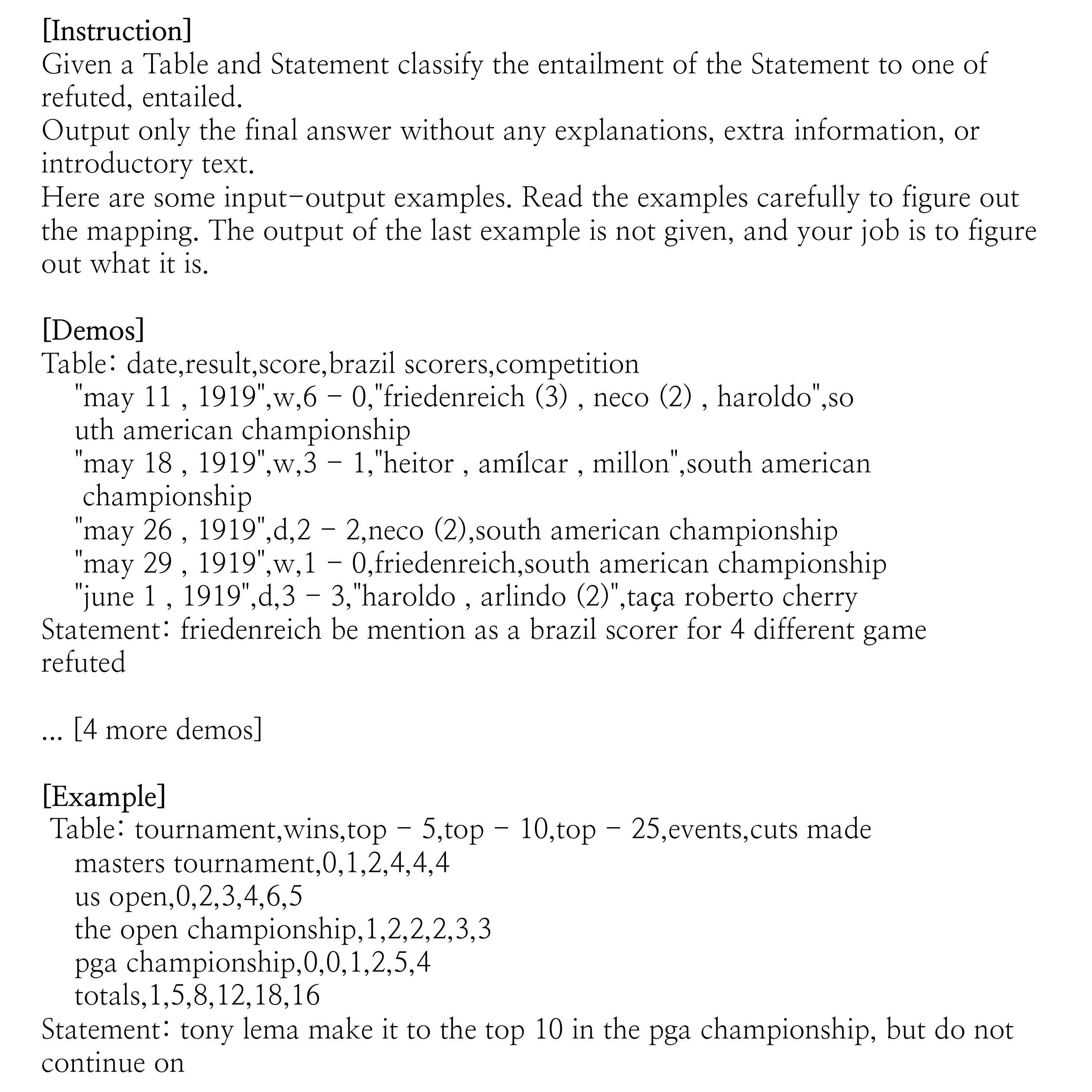}
    \caption{A prompt example from our benchmark, demonstrating the template and instructions used for \textit{TabFact}. The same prompt structure was applied across all datasets, with specific instructions tailored for each.}

    \label{fig:prompt_structure}
\end{figure*}

\begin{figure*}[b]
    \centering
    \includegraphics[width=\textwidth]{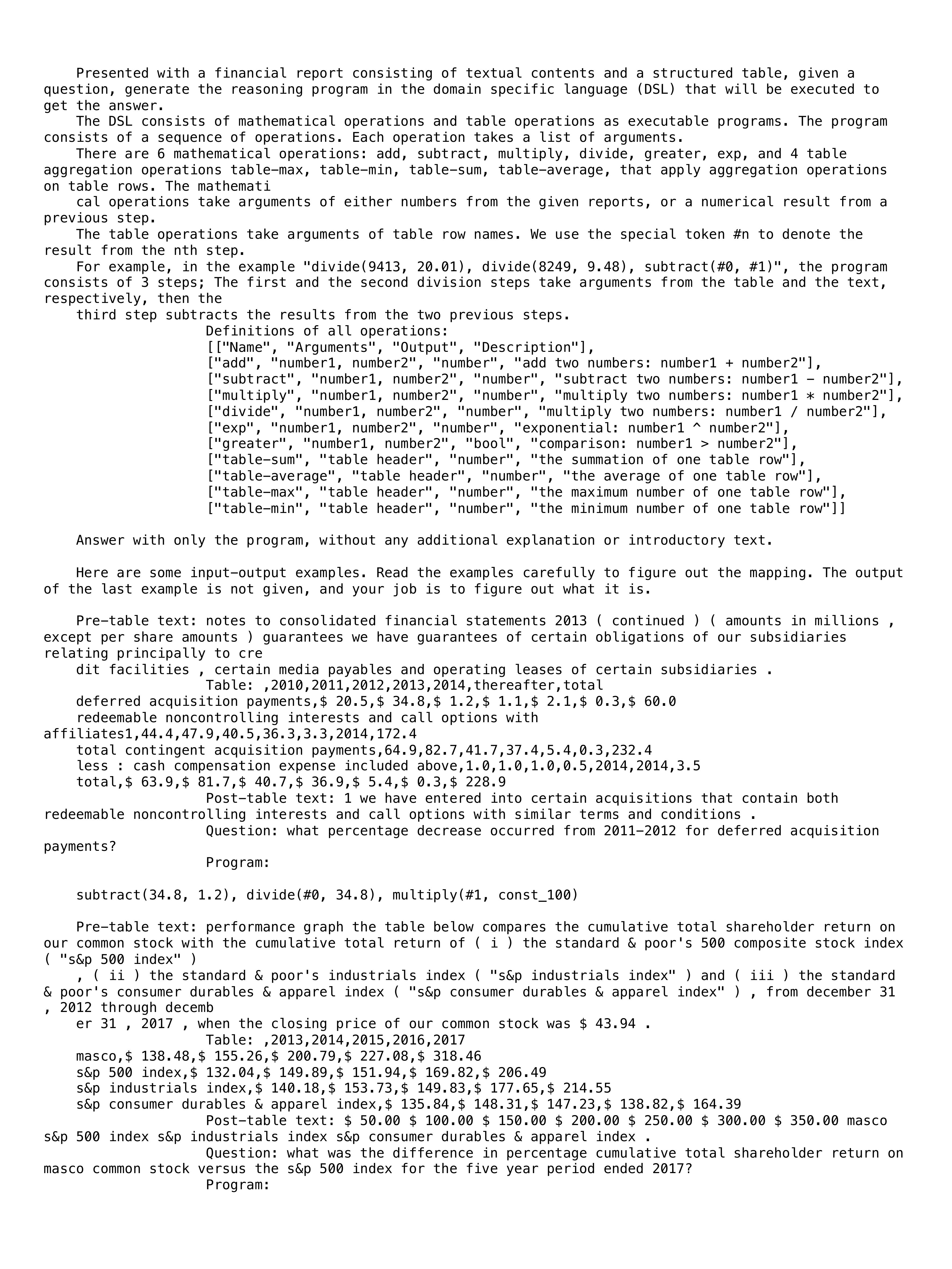}
    \caption{Example prompt used in FinQA. This single-shot prompt includes one demonstration that reflects both the input format and the expected output.}
    \label{fig:finqa_prompt}
\end{figure*}

\begin{figure*}
    \centering
    \includegraphics[width=\textwidth]{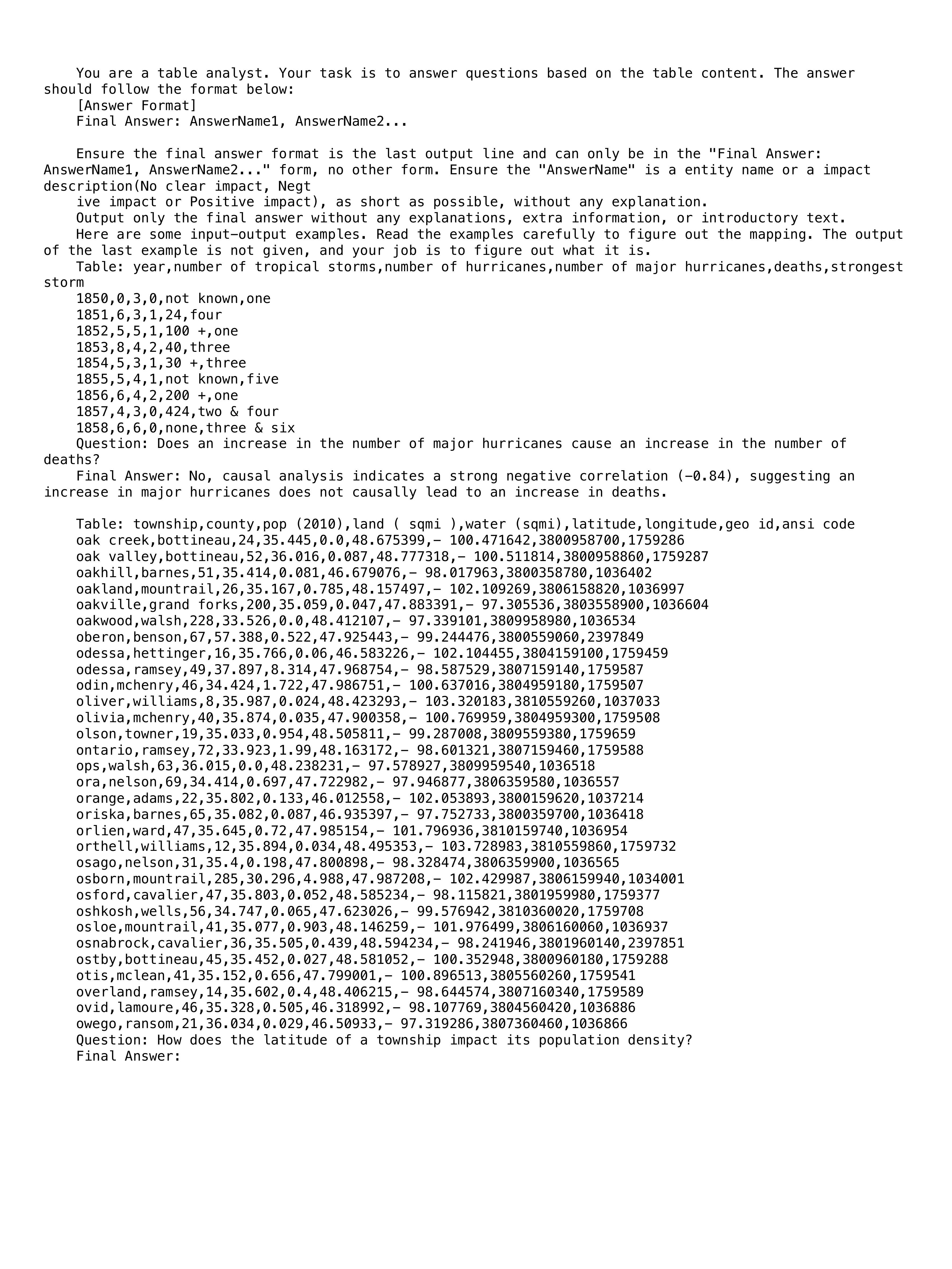}
    \caption{Example prompt used in TableBench datasets (DA, NR and FC). This single-shot prompt includes one demonstration that reflects both the input format and the expected output for TableBench DA.}
    \label{fig:tablebench_prompt}
\end{figure*}

\begin{figure*}
    \centering
    \includegraphics[width=\textwidth]{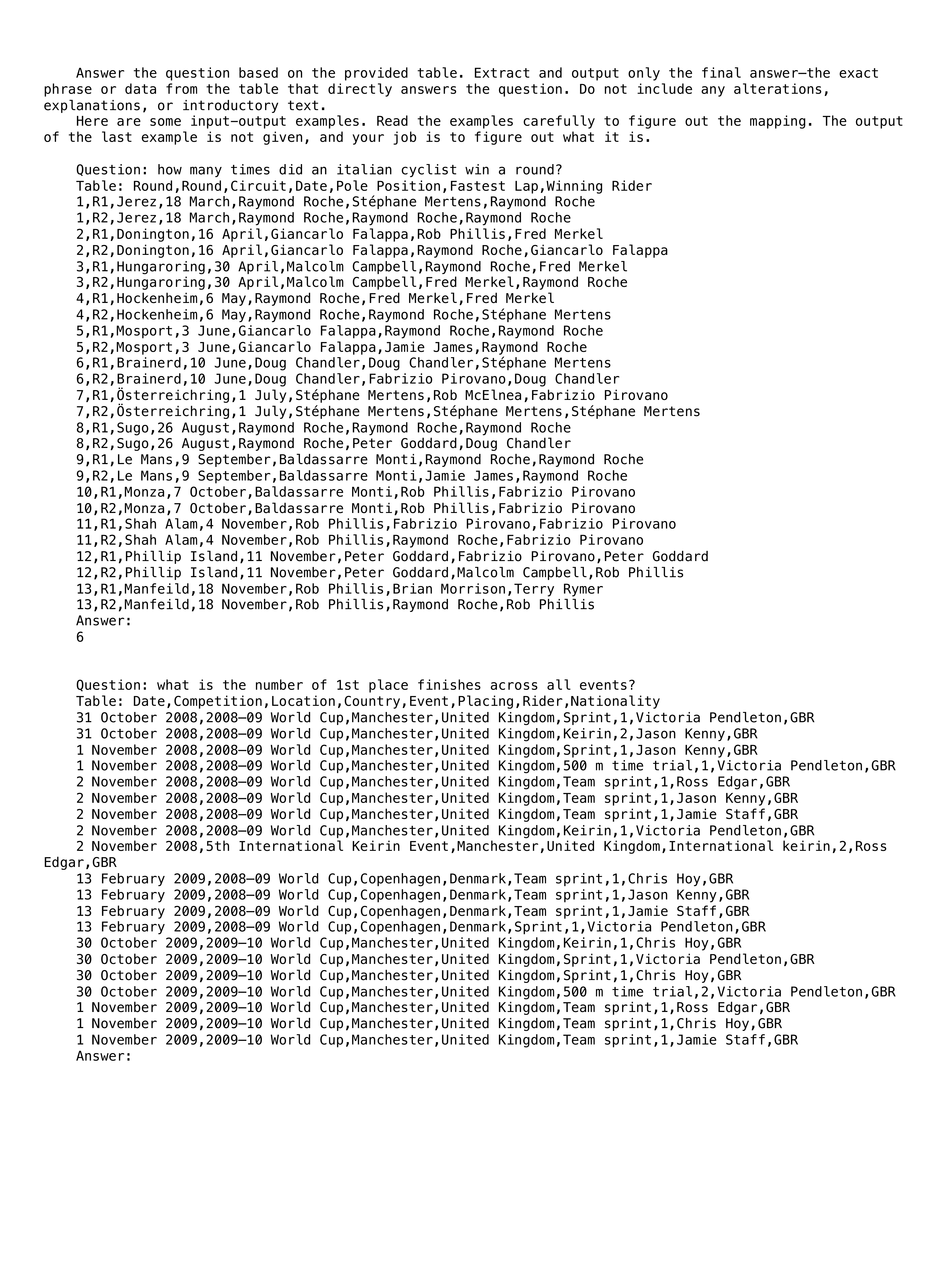}
    \caption{Example prompt used in WikiTQ. This single-shot prompt includes one demonstration that reflects both the input format and the expected output.}
    \label{fig:wikitq_prompt}
\end{figure*}

\begin{figure*}
    \centering
    \includegraphics[width=\textwidth]{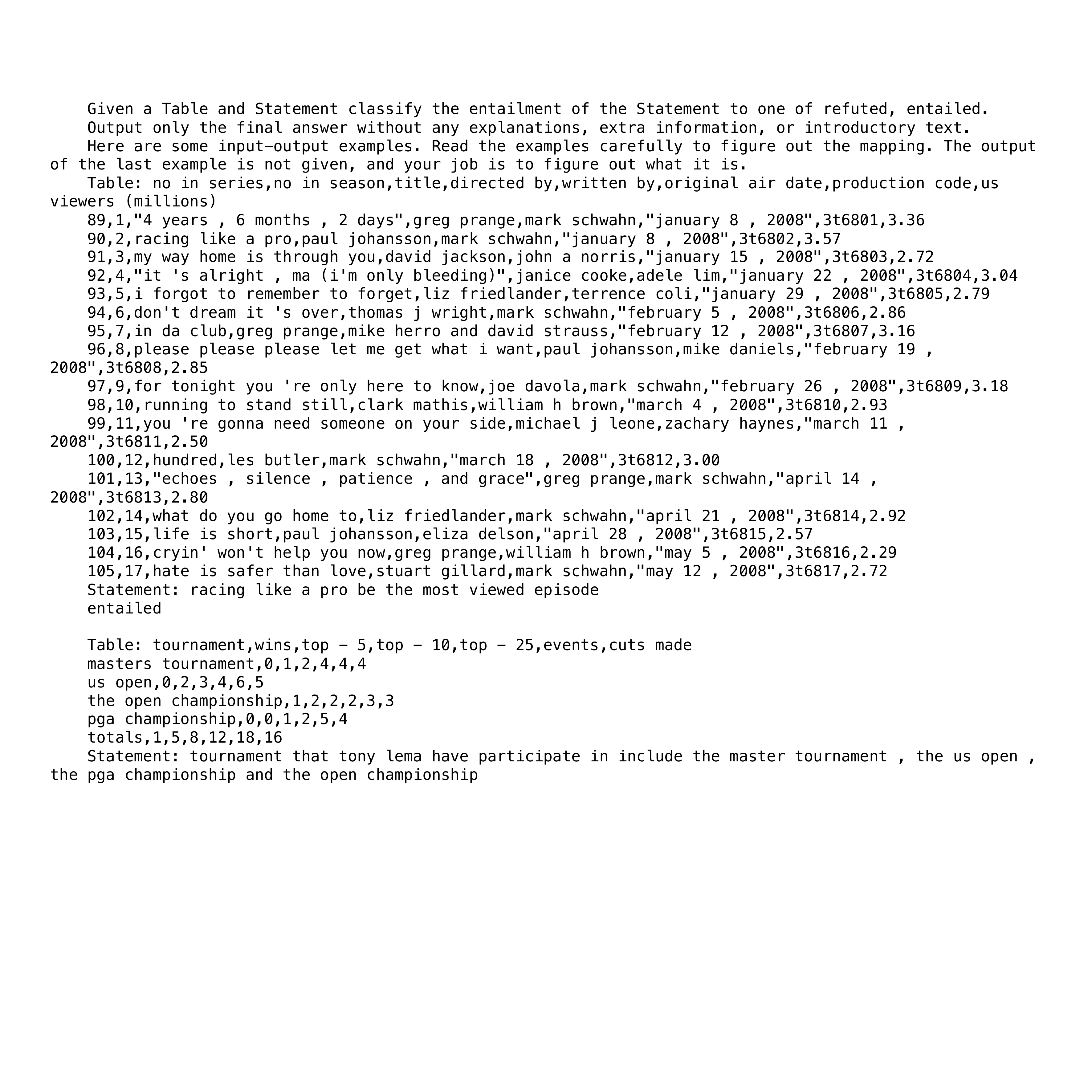}
    \caption{Example prompt used in TabFact. This single-shot prompt includes one demonstration that reflects both the input format and the expected output.}
    \label{fig:tabfact_prompt}
\end{figure*}

\begin{figure*}
    \centering
    \includegraphics[width=\textwidth]{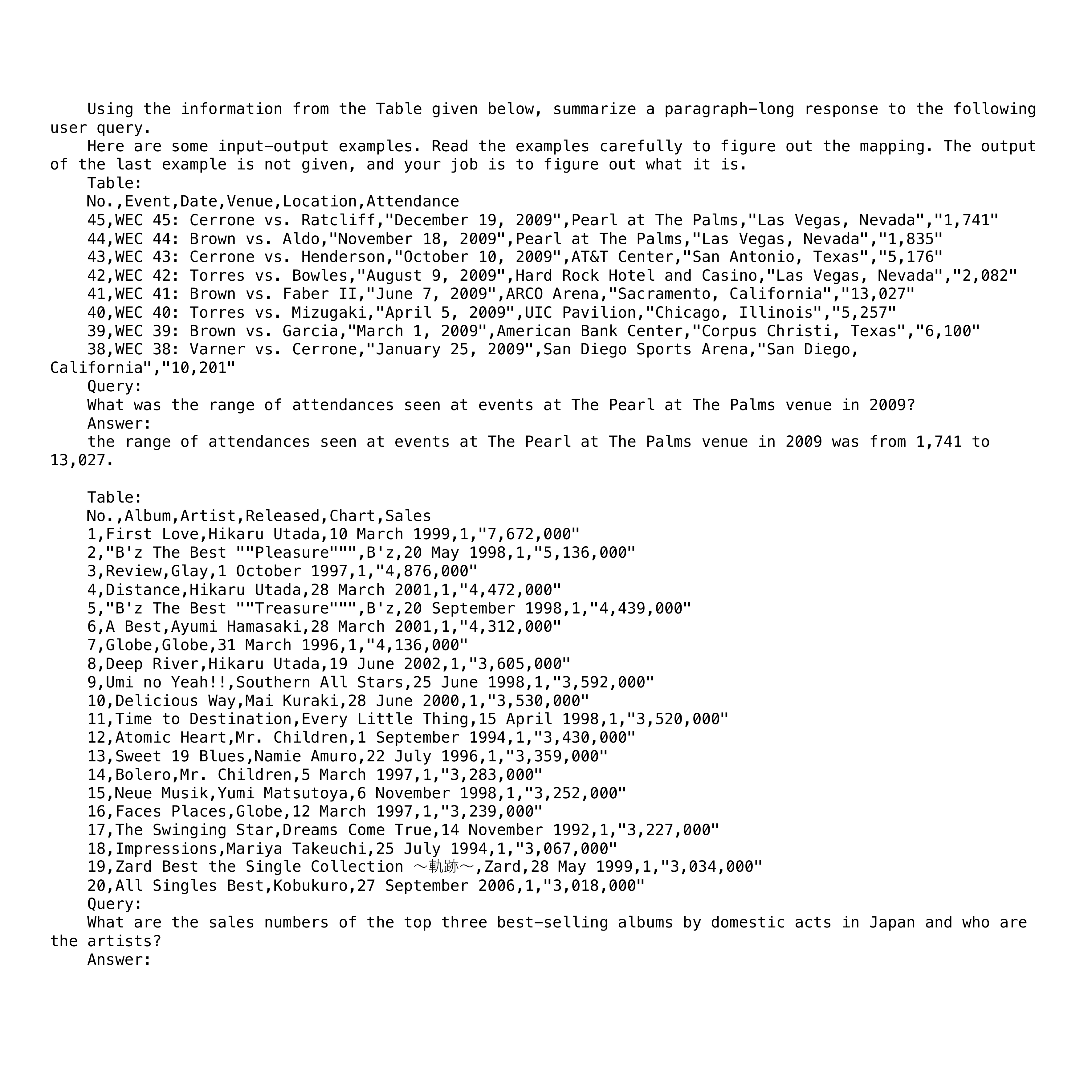}
    \caption{Example prompt used in QTSumm. This single-shot prompt includes one demonstration that reflects both the input format and the expected output.}
    \label{fig:qtsumm_prompt}
\end{figure*}

\begin{figure*}
    \centering
    \includegraphics[width=\textwidth]{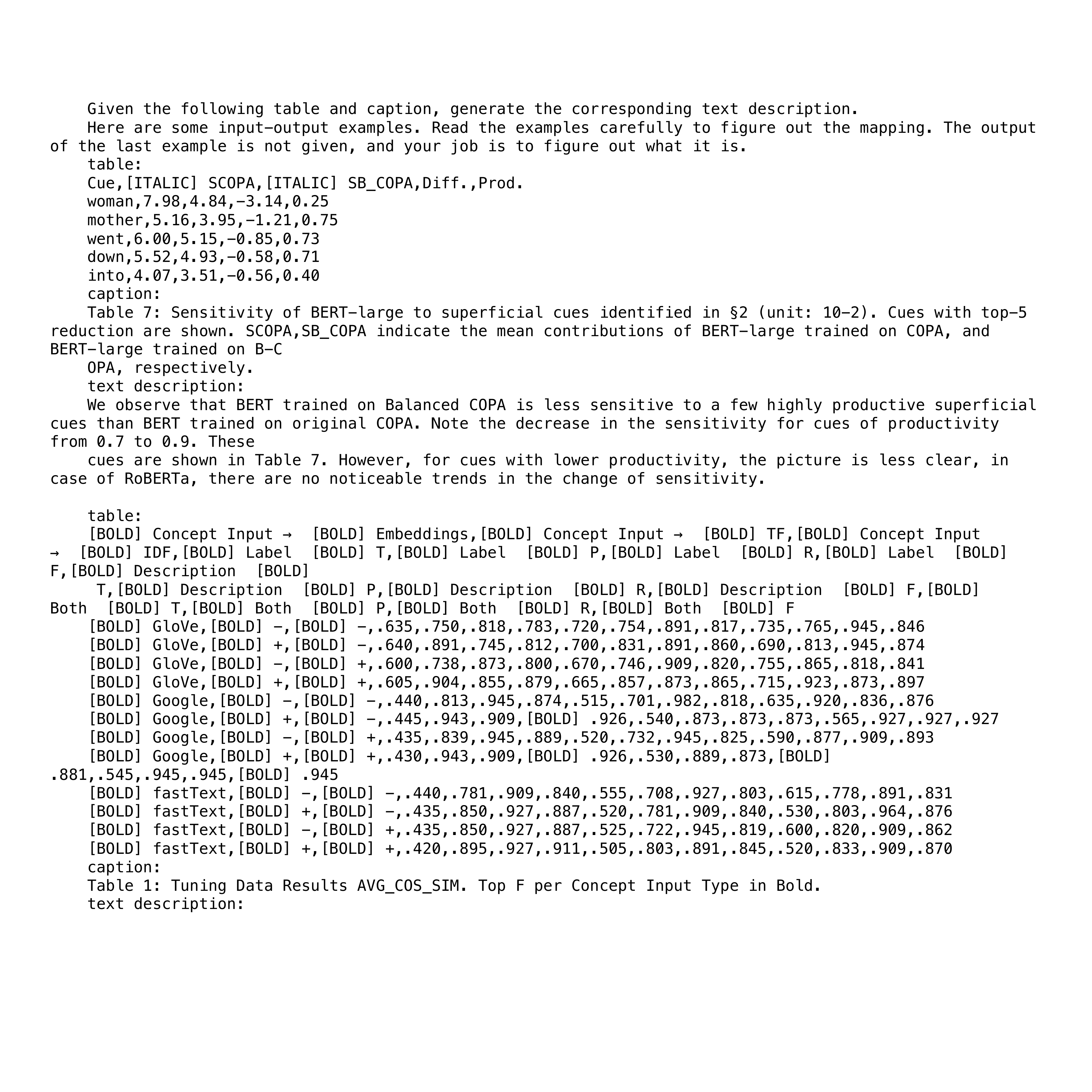}
    \caption{Example prompt used in Table-to-Text datasets (NumericNLG and SciGen). This single-shot prompt includes one demonstration that reflects both the input format and the expected output for SciGen.}
    \label{fig:scigen_prompt}
\end{figure*}

\begin{figure*}
    \centering
    \includegraphics[width=\textwidth]{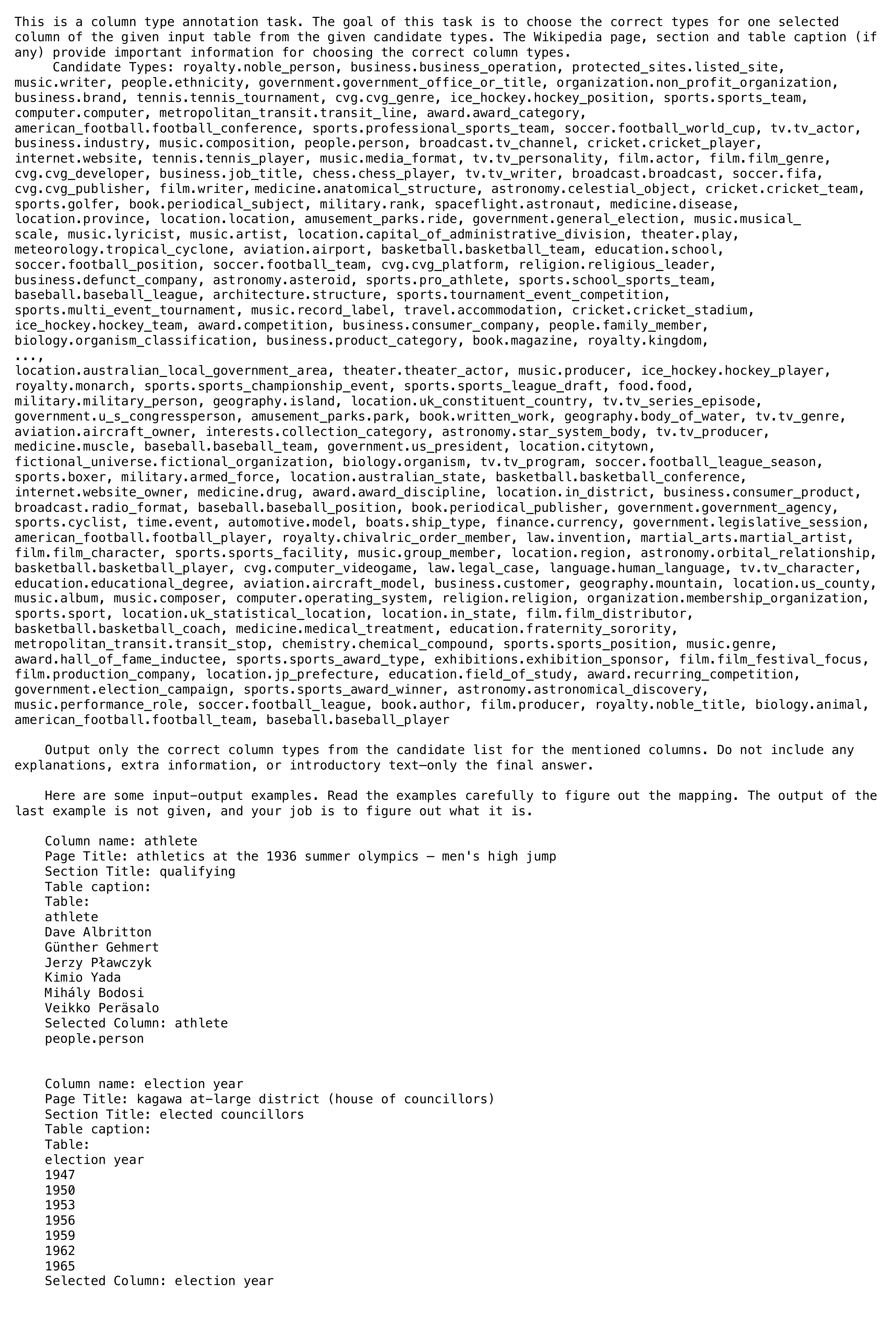}
    \caption{Example prompt used in TURL CTA. This single-shot prompt includes one demonstration that reflects both the input format and the expected output.}
    \label{fig:turl_prompt}
\end{figure*}

Here we provide examples of the prompts used in our benchmark. While the structure of each prompt includes the task instructions with highlights, five demonstrations with expected answers, and an input example (as illustrated in Figure~\ref{fig:prompt_structure}), we present them with a single demonstration for each task in Figures~\ref{fig:finqa_prompt}, \ref{fig:tablebench_prompt}, \ref{fig:wikitq_prompt}, \ref{fig:tabfact_prompt}, \ref{fig:qtsumm_prompt}, \ref{fig:scigen_prompt} and \ref{fig:turl_prompt}.


\end{document}